\PassOptionsToPackage{table}{xcolor}
\documentclass[10pt,twocolumn,letterpaper]{article}

\usepackage[pagenumbers]{cvpr}   %

\usepackage[utf8]{inputenc}
\usepackage[T1]{fontenc}
\usepackage{amsmath}
\usepackage{amsfonts}
\usepackage{graphicx}
\usepackage{booktabs}
\usepackage{multirow}
\usepackage{pifont}
\usepackage{wrapfig}
\usepackage[most]{tcolorbox}
\usepackage{xspace}

\definecolor{cvprblue}{rgb}{0.21,0.49,0.74}
\usepackage[pagebackref,breaklinks,colorlinks,allcolors=cvprblue]{hyperref}
\usepackage{etoc}
\definecolor{revblue}{RGB}{0,0,0}
\newcommand{\rev}[1]{{\color{revblue}#1}}
\newenvironment{revblock}{\color{revblue}}{}

\makeatletter
\renewcommand\paragraph{\@startsection{paragraph}{4}{\z@}%
  {1.5ex \@plus .3ex \@minus .2ex}{-1em}{\normalfont\normalsize\bfseries}}
\makeatother
\makeatletter
\def\cvprsect#1{\cvprsection[#1]{\texorpdfstring{\hskip -1em.~}{}#1}}
\def\cvprsubsect#1{\cvprsubsection[#1]{\texorpdfstring{\hskip -1em.~}{}#1}}
\def\cvprsubsubsect#1{\cvprsubsubsection[#1]{\texorpdfstring{\hskip -1em.~}{}#1}}
\makeatother

\def\confName{CVPR}
\def\confYear{2027}

\definecolor{boxgray}{RGB}{100, 100, 100}
\definecolor{mydarkred}{rgb}{0.6,0,0}
\definecolor{mydarkgreen}{rgb}{0,0.6,0}

\newtcolorbox{HeaderBox}[2][]{
    enhanced,
    title={#2},
    colframe=boxgray,
    colbacktitle=boxgray,
    colback=white,
    coltitle=white,
    fonttitle=\normalsize\bfseries,
    arc=2mm,
    boxrule=1.2pt,
    toptitle=1mm,
    bottomtitle=1mm,
    top=1mm,
    bottom=1mm,
    left=1mm,
    right=1mm,
    #1
}

\title{MarsRetrieval: Benchmarking Vision-Language Models \\ for Planetary-Scale Geospatial Retrieval on Mars}

\author{
  Shuoyuan Wang$^{1}$ \quad Yiran Wang$^{2}$ \quad Hongxin Wei$^{1}$\thanks{Corresponding author.} \\[4pt]
  $^{1}$Department of Statistics and Data Science, Southern University of Science and Technology \\
  $^{2}$Department of Earth and Space Sciences, Southern University of Science and Technology \\
  {\tt\small \{wangsy2024@mail.,wangyr@,weihx@\}sustech.edu.cn}
}

\begin{document}
\etocdepthtag.toc{mtbody}

\maketitle

\begin{abstract}
Data-driven approaches like deep learning are rapidly advancing planetary science, particularly in Mars exploration.
Despite recent progress, most existing benchmarks remain confined to closed-set supervised visual tasks and thus fail to support open-world retrieval in modern planetary exploration.
To bridge this gap, we introduce MarsRetrieval, a novel retrieval benchmark for evaluating the utility of vision-language models (VLMs) in Martian geospatial discovery.
MarsRetrieval adopts a unified retrieval-centric protocol to evaluate representative VLM architectures, including \rev{dual-encoder VLMs and MLLM-based embedding models}. 
Specifically, it organizes evaluation into three complementary tasks: (1) paired image-text retrieval, (2) landform retrieval, and (3) global geo-localization, spanning multiple spatial scales and diverse geomorphic features.
Through comprehensive evaluation, MarsRetrieval reveals a substantial challenge: even state-of-the-art VLMs (e.g., PE-Core \rev{and WeMM-Embedding}) struggle to distinguish fine-grained Martian geomorphic features in text-guided \rev{landform} retrieval.
Our results further demonstrate that domain-specific fine-tuning is critical for achieving robust and generalizable performance in planetary geospatial discovery.
By grounding the evaluation in real scientific challenges faced in Martian exploration, MarsRetrieval highlights the substantial gap between current multimodal AI capabilities and the practical needs of planetary research. Our code is available at \url{https://github.com/ml-stat-Sustech/MarsRetrieval}.
\end{abstract}

\section{Introduction}
\label{intro}

\begin{figure*}[t]
  \centering
  \includegraphics[width=0.98\textwidth]{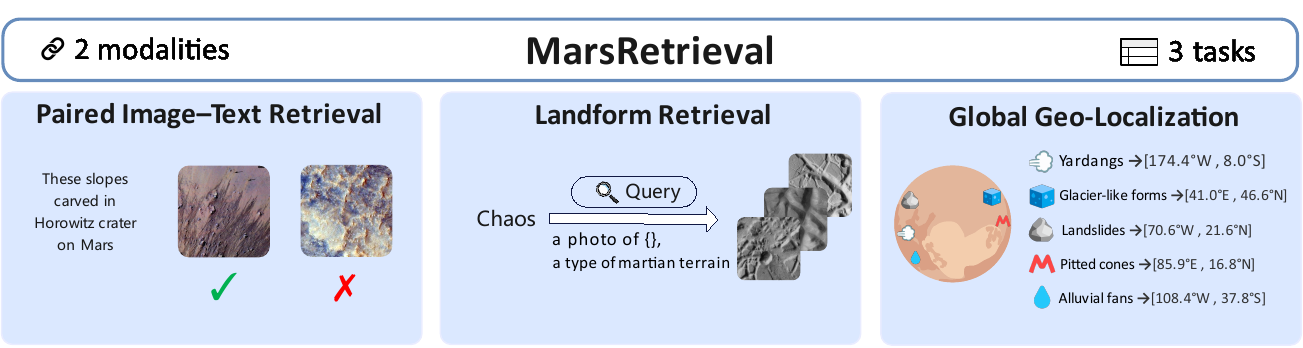}
  \caption{An overview of tasks in MarsRetrieval. The benchmark consists of 3 challenging tasks for cross-modal retrieval in Martian exploration. \rev{See Table~\ref{tab_marsretrieval_overview} for data sources and retrieval directions.}}
  \label{fig_marsretrieval_overview}
\end{figure*}

Planetary science plays a pivotal role in understanding the formation, evolution, and habitability of worlds beyond Earth.
Given its well-preserved record of planetary-scale processes (e.g., geological record, climate history, past aqueous activity), Mars serves as one of the most accessible and scientifically valuable targets in the Solar System~\cite{ehlmann2016sustainability}.
As high-resolution imagery from orbiters and rovers grows to unprecedented volumes, deep learning has emerged as a significant tool in various planetary science tasks, such as terrain classification~\cite{rothrock2016spoc, tiwari2024marsdeepnet}, landform mapping~\cite{wang2026natural, pearson2024mapping} and autonomous rover navigation~\cite{daftry2022mlnav, bechtold2023planetary}.
These advancements greatly accelerate scientific workflows and reduce human bias in interpretation.

The sustained progress in planetary science largely depends on high-quality benchmarks that facilitate reproducible evaluations across models.
A recent representative benchmark for planetary science is Mars-Bench~\cite{purohit2025marsbench}, which consolidates 20 Mars-related tasks covering image classification, segmentation, and detection. 
However, Mars-Bench primarily focuses on closed-set supervised visual tasks, which limits its applicability to real-world Martian exploration~\cite{fang2024mtsnet, wang2026natural}. 
In practice, planetary research often requires open-world, text-guided discovery and retrieval over large-scale geospatial data~\cite{parker2026aion, jakubik2025terramind}.
On the other hand, VLMs have established remarkable performance in various scientific domains, e.g., medical imaging~\cite{pan2025medvlm, zhang2023biomedclip}, chemistry~\cite{wang2025gtr, li2025chemvlm}, and earth observation~\cite{wang2025geollavak, kuckreja2024geochat}.
Although recent works have begun to introduce VLMs to planetary exploration, the community still lacks a standardized evaluation benchmark to assess their practical utility for Martian geospatial discovery.

To address this gap, we introduce MarsRetrieval, a novel and systematic retrieval benchmark for evaluating the utility of vision-language models in Martian geospatial discovery.
As shown in Figure~\ref{fig_marsretrieval_overview}, MarsRetrieval organizes evaluation into 3 complementary tasks spanning multiple spatial scales and diverse geomorphic features:
\textbf{(1) Paired Image-Text Retrieval} measures foundational vision-language alignment using curated Martian image-text pairs, covering diverse spatial scales \textemdash from global orbital mosaics to rover-level microscopic imagery.
\textbf{(2) Landform Retrieval} evaluates concept-to-instance generalization across 48 geomorphic categories organized by 7 major geological origins (e.g., Aeolian, Volcanic and Fluvial processes), supporting rapid scientific hypothesis verification at the concept level.
\textbf{(3) Global Geo-Localization} localizes scientific concepts within a global mosaic~\cite{dickson2023release, malin2007context} comprising over 1.4 million CTX tiles under massive background noise, facilitating planetary-scale applications such as global geomorphic cataloging.
All three tasks are built on \rev{expert-validated annotations or authoritative scientific catalogs}, ensuring scientific authority of the proposed benchmark.
In summary, by grounding these tasks in real scientific challenges faced in Martian exploration, MarsRetrieval provides a holistic benchmark for evaluating VLM practical utility across representative Martian scientific scenarios.

We establish strong baselines by benchmarking a broad spectrum of state-of-the-art foundation models under a unified retrieval-centric protocol. 
Our evaluation encompasses representative VLM architectures including \rev{dual-encoder VLMs~(e.g., CLIP~\cite{radford2021learning}), MLLM-based embedding models}~(e.g., Qwen3-VL-Embedding~\cite{li2026qwen3}) and vision-only foundation models~(e.g., DINOv3~\cite{simeoni2025dinov3}).
Results show that MarsRetrieval is highly challenging: \rev{MLLM-based embedding models are competitive with domain-specific models on paired image-text retrieval, but all general models lag far behind on landform retrieval, reaching at most 20.93 mAP compared with 71.89 for domain-specific models.}
We further find that domain-specific fine-tuning~\cite{wang2026natural} \rev{brings large gains, especially on landform retrieval.}
Overall, MarsRetrieval serves as a standardized evaluative framework for quantifying the reliability of VLMs for future Martian exploration and scientific analysis.
Furthermore, the proposed evaluation paradigm can be adapted to other autonomous planetary explorations, facilitating the development of foundation models for deep-space discovery.

\begin{table}[t]
\centering
\caption{\rev{Data sources and retrieval directions of the three tasks in MarsRetrieval.} We denote the modalities with $\text{I}$=image, $\text{T}$=text.}
\label{tab_marsretrieval_overview}
\resizebox{\columnwidth}{!}{
\begin{tabular}{llc}
\toprule
\textbf{Task} & \textbf{Data Sources} & \textbf{Modalities} \\
\midrule
Paired Image-Text Retrieval & Public + Curated & T $\to$ I; I $\to$ T \\
Landform Retrieval & CTX, HiRISE & T $\to$ I \\
Global Geo-Localization & CTX & T $\to$ I; I $\to$ I \\
\bottomrule
\end{tabular}
}
\end{table}

\section{Benchmark Overview}
Existing Mars-related benchmarks, such as Mars-Bench~\cite{purohit2025marsbench}, have established a standardized evaluation for planetary science. 
However, these benchmarks primarily focus on closed-set supervised vision tasks with single-modality inputs. 
This leaves a significant gap in evaluating the multimodal understanding of modern VLMs, which are essential for language-driven planetary research.
To bridge this gap, we introduce MarsRetrieval, a comprehensive retrieval-centric geospatial benchmark to evaluate diverse VLM embedding models on 3 Mars-related tasks.

\paragraph{Evaluation Protocol.} 
MarsRetrieval adopts a retrieval-based evaluation protocol, which evaluates whether matched items are ranked above mismatched candidates in a joint representation space.
Formally, given a query $q$ and a gallery of candidates $G = \{c_1, c_2, ..., c_n\}$, the model should rank the candidates based on their similarity $s(f(q), f(c_i))$, where $f(\cdot)$ represents the model's embedding function and $s(\cdot, \cdot)$ is typically the cosine similarity.
By utilizing this protocol, MarsRetrieval can objectively measure zero-shot understanding of VLMs across diverse tasks without the need for task-specific fine-tuning.

\paragraph{Design Principles.}
Under the unified retrieval-centric protocol mentioned above, MarsRetrieval proposes three complementary tasks that form a comprehensive evaluation: from foundational perception to planetary-scale application. 
\rev{As illustrated in Figure~\ref{fig_marsretrieval_overview} and summarized in Table~\ref{tab_marsretrieval_overview},} these tasks are designed to evaluate distinct capability dimensions: 
(1) Paired Image-Text Retrieval assesses foundational feature alignment by matching fine-grained scientific descriptions with corresponding Martian imagery across diverse orders of spatial scale. 
(2) Landform Retrieval evaluates concept-to-instance generalization by retrieving diverse instances from 48 long-tailed geomorphic categories spanning multiple geological origins (e.g., aeolian, volcanic) using their scientific concepts.
(3) Global Geo-Localization simulates planetary-scale discovery by localizing target landforms within a global archive of 1.4 million tiles under massive background noise.
We describe three tasks and their datasets and evaluation protocols in Sections~\ref{task_1}, \ref{task_2}, and \ref{task_3}, respectively.

\section{Task 1: Paired Image-Text Retrieval}
\label{task_1}
\begin{figure}[t]
    \centering
    \includegraphics[width=\linewidth]{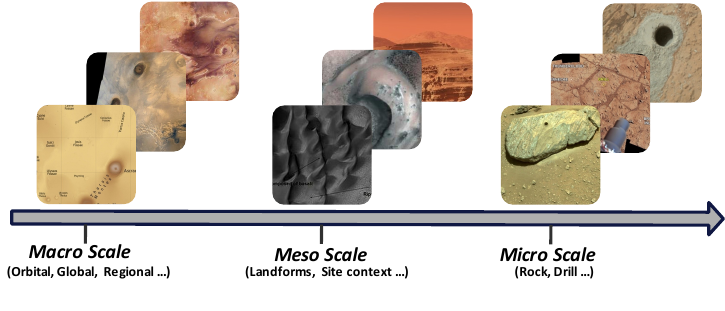}
    \caption{Planetary scale coverage in task 1 (Paired Image-Text Retrieval). We include Martian paired vision-language samples from orbital terrain to rover instrumentation.}
    \label{fig_task1_data}
\end{figure}
We construct the dataset via a multi-stage pipeline that prioritizes Mars relevance, pair quality, and broad coverage.
We formulate task 1 as a one-to-one cross-modal retrieval task, where a model should align highly specialized planetary-science descriptions with the corresponding Martian image (and vice-versa).
The curated dataset consists of 2,287 Martian paired samples, spanning diverse orders of magnitude in spatial scale.
As shown in Figure~\ref{fig_task1_data}, the dataset coverage ranges from planetary-scale orbital mosaics to rover-level microscopic imagery, encompassing diverse imaging modalities, geomorphic structures, atmospheric phenomena, and instrument-related visual cues.

\subsection{Dataset Construction}
\label{task1_data_operation}

\paragraph{Data Collection.}
\rev{We build a candidate pool of Martian image-text pairs from two web-scale corpora: DataComp-1B~\cite{gadre2023datacomp} and Re-LAION-2B-en~\cite{schuhmann2022laion} via caption-based retrieval with FAISS~\cite{douze2025faiss} and Qwen3-Embedding-0.6B~\cite{zhang2025qwen3}.
The query set covers major Mars-related concepts: geomorphic processes and landforms, atmospheric phenomena, and multi-scale remote-sensing data (e.g., CTX~\cite{dickson2023release, malin2007context}, HiRISE~\cite{mcewen2007mars, mcewen2024high}).}

\paragraph{Data Filtering.}
\rev{To ensure pair correctness and Mars specificity, we retain a pair only if it passes the thresholds of three complementary automatic scores:} (1) \textbf{CLIP score:} \rev{CLIPScore~\cite{hessel2021clipscore} with ViT-H-14-378-quickgelu measures basic cross-modal similarity.} (2) \textbf{LLM relevance confidence:} \rev{Qwen2.5-7B~\cite{qwen2025qwen25technicalreport} verifies whether the caption refers to Mars-related concepts via token-level confidence~\cite{luo2025your}.} (3) \textbf{MLLM consistency confidence:} \rev{Qwen2.5-VL-7B-Instruct~\cite{bai2025qwen2} assesses whether the visual evidence is grounded in the text and relevant to Mars. Appendix~\ref{appx_task1_data} details the filtering strategy, data statistics and threshold analysis, with prompts in Figures~\ref{prompt_llm_filtering} and~\ref{prompt_mllm_filtering}.}

\paragraph{Diversity Sampling.}
To improve coverage of scientific concepts and visual appearances, we select a compact yet diverse subset via a two-stage K-center Greedy strategy.
First, we select a subset that maximizes semantic diversity in the text embedding space using Qwen3-Embedding-0.6B~\cite{zhang2025qwen3}.
Second, within the subset, we further maximize coverage in the visual feature space using DINOv3~\cite{simeoni2025dinov3}.

\paragraph{Caption Refinement.}
The original web-scale captions may be noisy (e.g., internal IDs, instrument metadata, non-descriptive titles).
Hence, we utilized Seed2.0-Lite\rev{~\cite{seed2026seed2}} to write a visual caption draft for each image, and Seed2.0-Mini to fuse the draft with the original caption into a final caption.
We rewrite more precise captions while remaining strictly grounded in visible evidence. The effectiveness of caption refinement is provided in Figure~\ref{fig_task1_ablation_caption}.

\paragraph{Expert Validation.}
To guarantee scientific authority, we adopt a two-stage human annotation protocol.
In the first stage, three junior researchers with planetary geomorphology backgrounds independently annotate each sample, and we retain only the intersection of their annotations.
\rev{The four annotators (one professor, one PhD student and two master's students) all have planetary-science research backgrounds, are familiar with Martian geomorphology and have complementary research specializations.}
We assess inter-annotator agreement using Fleiss' $\kappa$~\cite{fleiss1971measuring}, obtaining $\kappa=0.782$ and unanimous consensus on $96.36\%$ ($2409/2500$) of samples.
In the second stage, a senior planetary scientist reviews the consensus set and performs final adjudication to ensure Mars relevance and scientific correctness, validating $99.65\%$ ($2287/2295$) of the vision-language pairs.
The statistics confirm that junior-level annotations are well-aligned with expert judgment.

\subsection{Evaluation Set-up and Metrics}

{\setlength{\parindent}{0pt}%
\paragraph{\bfseries\upshape Evaluation Setup.}
We evaluate the model on bidirectional image-text retrieval. 
All image-text pairs are encoded into a shared embedding space and the embeddings are L2-normalized. 
Given a query in one modality (e.g., Image), candidates in the other modality (e.g., Text) are ranked by cosine similarity. The rank of the paired ground-truth instance is recorded to calculate retrieval metrics.
}

{\setlength{\parindent}{0pt}%
\paragraph{\bfseries\upshape Metrics.}
We evaluate performance using standard retrieval metrics \textbf{Recall@1 (R@1)} and \textbf{Recall@10 (R@10)}, which measure the hit rate of the ground-truth match within the top-$K$ ranked candidates.
In addition, to better characterize ranking quality beyond a fixed cutoff, we report \textbf{Mean Reciprocal Rank (MRR)}, which rewards placing the true match as high as possible.
Finally, we report \textbf{Median Rank (MedR)}, \rev{which is robust to outliers (lower is better).} Formal definitions are shown in Appendix~\ref{appx_task1_metrics}.
}

\section{Task 2: Landform Retrieval}
\label{task_2}
\begin{figure}[t]
\centering
\includegraphics[width=\linewidth]{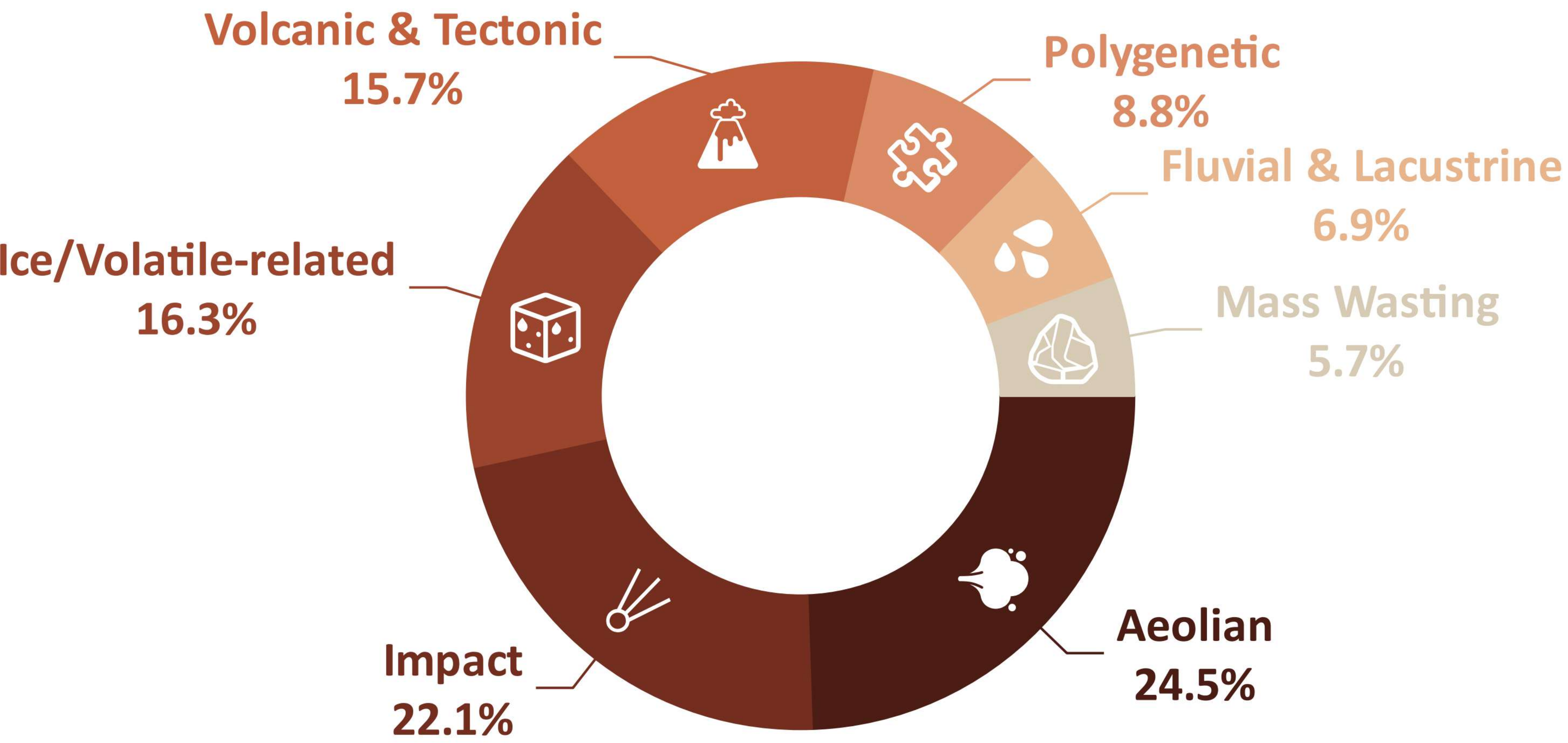}
\caption{Dataset distribution across the 7 major genetic classes of Martian landforms in Task 2 (Landform Retrieval). We provided the detailed sample distribution in Figure~\ref{fig_task2_detailed_distribution}.}
\label{fig_task2_data}
\end{figure}

We formulate Task 2 as a text-to-image retrieval task.
Given a textual prompt of a geomorphic concept, the model should retrieve its multiple corresponding visual instances from a Martian gallery.
The dataset comprises 1,185 carefully curated image patches and follows a two-level geomorphology taxonomy.
As shown in Figure~\ref{fig_task2_data}, we organize the dataset into 7 major genetic classes.
The sub-classes are the actual retrieval units, which consist of 48 landforms and follow a long-tailed distribution (See Figure~\ref{fig_task2_detailed_distribution} in Appendix).
A key challenge is that Martian landforms do not conform to a single canonical appearance.
Specifically, Task~2 emphasizes 2 challenges: (1) Morphological diversity (e.g., different erosion stages, slopes, scales), (2) Spatial diversity (e.g., varying latitudes and illumination).

\subsection{Dataset Construction}

\rev{Task~2 tests whether a model remains robust across variations in morphology and geography.
To this end, we combine foundation model discovery with curation from CTX and HiRISE imagery.}

{\setlength{\parindent}{0pt}%
\paragraph{\bfseries\upshape Data Collection.}
\rev{Following the retrieval pipeline of MarScope~\cite{wang2026natural}, we perform} large-scale text-to-image similarity search across a global Mars mosaic at default scale.
\rev{To mitigate the bias of a single model, we use multiple MarScope checkpoints trained with different configurations.}
For each landform subclass, we retrieve the top-100 candidates based on cosine similarity.
Moreover, to cover rare classes that are difficult to observe reliably at the default observation scale of MarScope  (e.g., Boulder Track, Fractured Mounds, etc.), we supplement the dataset with targeted high-resolution image patches curated from public-domain CTX~\cite{dickson2024global} and HiRISE~\cite{mcewen2007mars, mcewen2024high} data.
To ensure the fairness of our evaluation, we further provide a detailed analysis of model generalization on these curated classes without using MarScope in \rev{Table}~\ref{tab_task2_marscope_bias}.
}

{\setlength{\parindent}{0pt}%
\paragraph{\bfseries\upshape Expert Validation.}
We adopt the same two-stage human annotation protocol as in Task~1 (Section~\ref{task1_data_operation}).
Planetary geomorphology experts manually verify the collected candidates to ensure each retained image exhibits discriminative characteristics consistent with its target landform.
Crucially, we preserve intra-class diversity by retaining samples from different latitudes and morphologies.
}

\subsection{Evaluation Set-up and Metrics}

\label{task2_setup}

{\setlength{\parindent}{0pt}%
\paragraph{\bfseries\upshape Evaluation Setup.}
We evaluate text-to-image retrieval under a multi-positive setting.
For each sub-class, we generate a prototype query embedding averaged from prompt templates using the model's text encoder \rev{(details in Appendix~\ref{appx_task2_prompt_ensemble}, ablation in Table~\ref{tab_task2_prompt_ablation}).}
The visual database is pre-indexed offline using the model's image encoder. 
During retrieval, all database images are ranked by cosine similarity, and the ground-truth positives are all images annotated with the corresponding subclass.
}

{\setlength{\parindent}{0pt}%
\paragraph{\bfseries\upshape Metrics.}
We report metrics for multi-positive retrieval under long-tail class frequencies.
Our primary metric is \textbf{macro-mAP}, which computes Average Precision per sub-class and then averages across concepts.
We also report Normalized Discounted Cumulative Gain \textbf{nDCG@10} to assess whether representative positives are prioritized early in the ranked list.
Finally, we report \textbf{Hits@10} to measure whether at least one true instance appears in the top-10 results.
\rev{Formal definitions are in Appendix~\ref{appx_task2_metrics}.}}

\section{Task 3: Global Geo-Localization}
\label{task_3}

\begin{figure}[t]
\centering
\includegraphics[width=\linewidth]{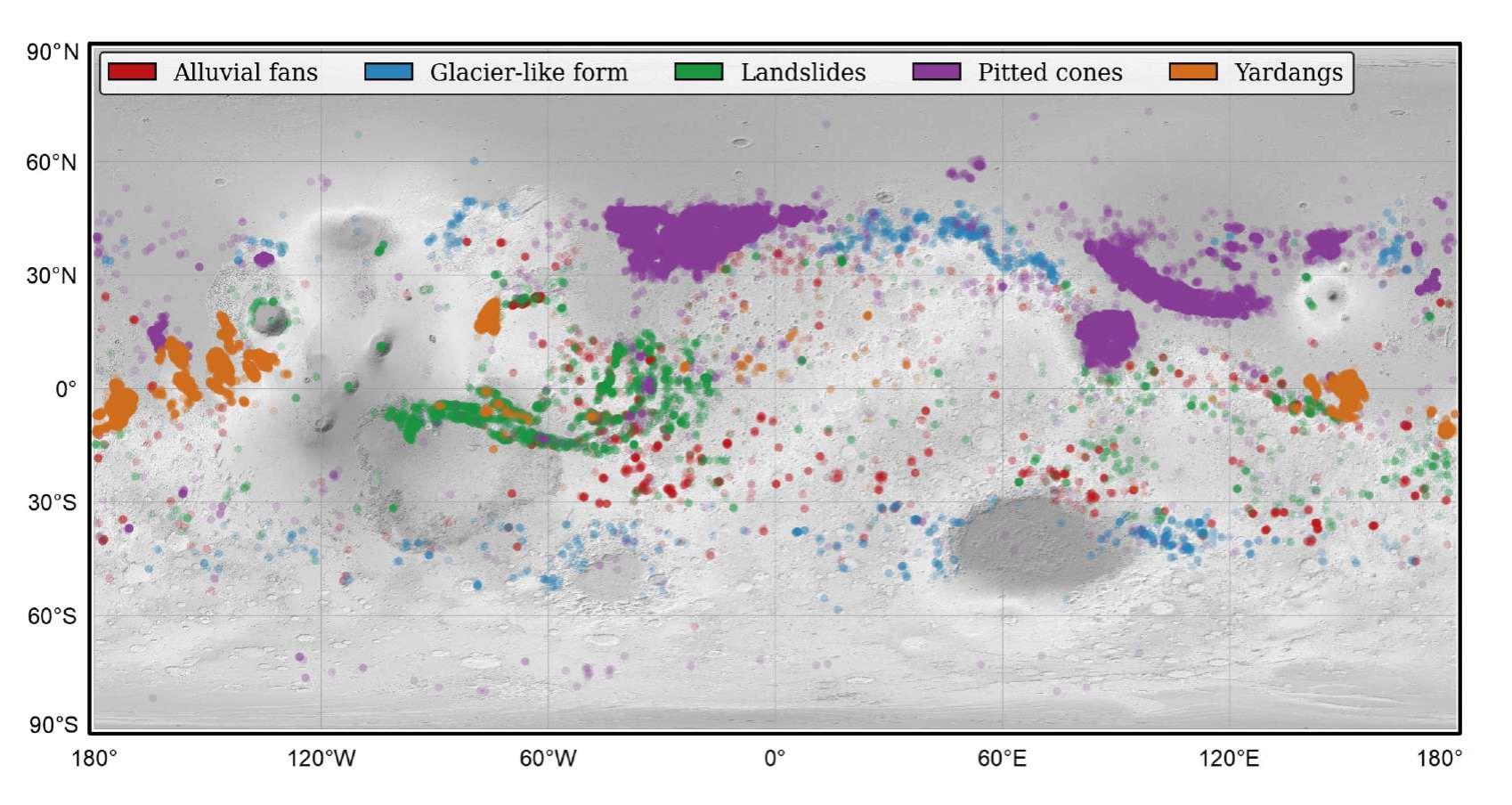}
\caption{Global ground-truth distribution from scientific catalogs~\cite{morgan2022global,souness2012inventory,roback2021controls,mills2024global,liu2020mapping} for Task 3. Different colors denote five target landforms, showing their distributions \rev{across Mars}.}
\label{fig_task3_data}
\end{figure}

We formulate Task~3 as a planetary-scale geo-localization problem.
Given a query (text or image), the model is expected to retrieve all relevant geographic locations from the global CTX mosaic~\cite{dickson2023release, malin2007context}.
The set of predicted coordinates could further form a coherent global distribution map on the Mars surface.
Unlike Tasks~1\&2, which operate on a closed candidate gallery, Task~3 emphasizes real-world deployment under extreme class imbalance: sparse positives embedded in massive background noise.
\rev{Figure~\ref{fig_task3_data} shows the global distributions of the five landforms from published scientific catalogs.}

\subsection{Dataset Construction}

\paragraph{Global Mosaic Tiling.}
\rev{Following the tiling protocol of MarScope~\cite{wang2026natural},} we construct a planetary-scale retrieval index by partitioning the global Mars CTX mosaic (\rev{$\sim$}6 m/pixel)~\cite{dickson2024global} into a large number of overlapping tiles.
Specifically, we adopt a default angular resolution of $0.2^\circ$ (approximately 12 km at the equator) to generate a global gallery comprising over 1.4 million patches.
Each patch is indexed with precise planetocentric coordinates via spherical projection. 
\rev{MarScope is used here only for this tiling procedure: it is not used to select gallery tiles or ground-truth locations, and the five catalogues below are not used to train MarScope.}

\paragraph{Ground Truth Curation.}
\rev{We aggregate ground-truth locations (planetocentric latitude and longitude) from published global catalogs of Martian landforms:} Alluvial fans~\cite{morgan2022global}, Glacier-like forms~\cite{souness2012inventory}, Landslides~\cite{roback2021controls}, Pitted cones~\cite{mills2024global}, and Yardangs~\cite{liu2020mapping}.
\rev{All five catalogues are manually annotated or verified by domain experts.}
\rev{These catalogues are the most authoritative global references currently available;} for example, the alluvial fan database~\cite{morgan2022global} and glacier-like form inventory~\cite{souness2012inventory} have served as standard references across numerous geomorphic and climatic studies of Mars~\cite{wilson2021global, brough2016former, hepburn2020polyphase, driver2024large}.
\rev{Task~3 is restricted to these five landforms because global geo-localization requires an authoritative, planet-wide catalog as ground truth: for landforms with ambiguous definitions or incomplete global inventories, valid but uncatalogued detections would be counted as false positives. Task~2 avoids this constraint, since experts validate a finite set of collected patches rather than every occurrence on Mars.}

\subsection{Evaluation Set-up and Metrics}
\label{task3_setup}

{\setlength{\parindent}{0pt}%
\paragraph{\bfseries\upshape Evaluation Setup.}
We evaluate two query modes: text-based and image-based.
The prompt template and image queries are shown in Appendix~\ref{appx_task2_prompt_ensemble} and Appendix~\ref{appx_task3_data}, respectively.
The image embeddings are extracted using the model's visual encoder and stored by FAISS in advance.
For each target landform, we retrieve the top-$K$ (e.g., $K=20,000$) candidates by cosine similarity \rev{and map them to a predicted point set.}

For quantitative validation, we evaluate retrieval outputs against published global landform catalogues as point sets.
We formulate evaluation as a proximity-based point-set matching problem between the prediction set and the catalogue reference set.
Since exact coordinate coincidence is infeasible due to continuous geography, we consider an image patch to be a true positive if the predicted coordinate falls within a physical spatial tolerance radius $r$ (e.g., $r=0.5^\circ$) of a ground-truth entry. We conduct the approximate calculation via KDTree~\cite{bentley1975multidimensional}.
The choice of $r=0.5^\circ$ corresponds to approximately 2.5 tile widths ($\sim$30~km at the equator), which is consistent with the typical spatial extent of the target landforms (e.g., Alluvial Fans 10--40~km, Landslides 5--50~km). We provide a detailed ablation on the tolerance radius in Figure~\ref{fig_task3_ablation_r}.
}

{\setlength{\parindent}{0pt}%
\paragraph{\bfseries\upshape Metrics.}
Given the extreme sparsity of positive samples against the global background, standard accuracy metrics are unsuitable.
We report \textbf{AUPRC (Area Under Precision-Recall Curve)} as the primary metric, which \rev{assesses the precision-recall trade-off} without being biased by the massive volume of true negatives.
Additionally, we report the optimal F1 score \textbf{$\text{F1}@{K^\star}$} on the PR curve \rev{to quantify the best alignment between the predicted map and the catalogue.} \rev{Formal definitions are in Appendix~\ref{appx_task3_metrics}.}
}

\begin{table*}[t]
\caption{Retrieval results of Task~1 (Paired Image-Text Retrieval) using Seed2.0-refined captions. We report bidirectional retrieval performance on various vision-language models, grouped by model scale. \rev{Best result per column within each model group is in bold. \rev{Bootstrap 95\% confidence intervals are reported in Appendix~\ref{appx_disc_ci}.}}}
\centering
\resizebox{\textwidth}{!}{
\renewcommand\arraystretch{1.15}
\begin{tabular}{cll cccc cccc}
\toprule
\multirow{2}{*}{\textbf{Size}} & \multirow{2}{*}{\textbf{Model}} & \multirow{2}{*}{\textbf{Backbone}} & \multicolumn{4}{c}{\textbf{T$\to$I}} & \multicolumn{4}{c}{\textbf{I$\to$T}} \\
\cmidrule(lr){4-7}\cmidrule(lr){8-11}
 &   &   & \textbf{R@1} & \textbf{R@10} & \textbf{MRR} & \textbf{MedR} & \textbf{R@1} & \textbf{R@10} & \textbf{MRR} & \textbf{MedR} \\
\midrule
\rowcolor{gray!25}\multicolumn{11}{c}{\textbf{\textit{Encoder-based Models}}} \\
\midrule
\color{revblue}\multirow{6}{*}{0.4B} & \color{revblue}DFN2B-CLIP-ViT-L-14 & \color{revblue}ViT-L/14 & \color{revblue}17.32 & \color{revblue}42.50 & \color{revblue}25.67 & \color{revblue}17 & \color{revblue}20.94 & \color{revblue}47.57 & \color{revblue}29.87 & \color{revblue}12 \\
  & \color{revblue}ViT-L-16-SigLIP-384 & \color{revblue}ViT-L/16 & \color{revblue}21.60 & \color{revblue}42.94 & \color{revblue}29.20 & \color{revblue}18 & \color{revblue}21.73 & \color{revblue}45.34 & \color{revblue}29.78 & \color{revblue}15 \\
  & \color{revblue}aimv2-large-patch14-224 & \color{revblue}ViT-L/14 & \color{revblue}10.06 & \color{revblue}29.43 & \color{revblue}16.83 & \color{revblue}43 & \color{revblue}11.76 & \color{revblue}33.89 & \color{revblue}19.15 & \color{revblue}34 \\
  & \color{revblue}ViT-L-16-SigLIP2-512 & \color{revblue}ViT-L/16 & \color{revblue}25.89 & \color{revblue}50.33 & \color{revblue}34.06 & \color{revblue}10 & \color{revblue}29.38 & \color{revblue}55.27 & \color{revblue}37.95 & \color{revblue}7 \\
  & \color{revblue}PE-Core-L-14-336 & \color{revblue}ViT-L/14 & \color{revblue}27.11 & \color{revblue}57.37 & \color{revblue}37.51 & \color{revblue}6 & \color{revblue}28.99 & \color{revblue}59.64 & \color{revblue}38.76 & \color{revblue}6 \\
  & \color{revblue}MarScope & \color{revblue}ViT-L/14 & \color{revblue}\textbf{35.24} & \color{revblue}\textbf{72.02} & \color{revblue}\textbf{47.70} & \color{revblue}\textbf{3} & \color{revblue}\textbf{34.54} & \color{revblue}\textbf{70.14} & \color{revblue}\textbf{46.57} & \color{revblue}\textbf{3} \\
\midrule
\rowcolor{gray!25}\multicolumn{11}{c}{\textbf{\textit{MLLM-based Models}}} \\
\midrule
\color{revblue}\multirow{7}{*}{$\leq$3B} & \color{revblue}jina-embeddings-v4 & \color{revblue}Qwen2.5-VL & \color{revblue}17.32 & \color{revblue}33.80 & \color{revblue}22.98 & \color{revblue}57 & \color{revblue}17.14 & \color{revblue}34.63 & \color{revblue}23.08 & \color{revblue}46 \\
  & \color{revblue}Ops-MM-embedding-v1-2B & \color{revblue}Qwen2-VL & \color{revblue}20.99 & \color{revblue}40.14 & \color{revblue}27.68 & \color{revblue}24 & \color{revblue}22.08 & \color{revblue}39.88 & \color{revblue}28.40 & \color{revblue}27 \\
  & \color{revblue}Qwen3-VL-Embedding-2B & \color{revblue}Qwen3-VL & \color{revblue}22.30 & \color{revblue}44.38 & \color{revblue}29.85 & \color{revblue}16 & \color{revblue}21.91 & \color{revblue}42.15 & \color{revblue}29.03 & \color{revblue}23 \\
  & \color{revblue}UEmbed-2B & \color{revblue}Qwen3.5 & \color{revblue}26.72 & \color{revblue}50.98 & \color{revblue}34.90 & \color{revblue}10 & \color{revblue}25.36 & \color{revblue}46.74 & \color{revblue}32.50 & \color{revblue}14 \\
  & \color{revblue}WeMM-Embedding-2B & \color{revblue}Qwen3.5 & \color{revblue}36.38 & \color{revblue}62.61 & \color{revblue}45.43 & \color{revblue}4 & \color{revblue}33.23 & \color{revblue}61.08 & \color{revblue}42.72 & \color{revblue}5 \\
  & \color{revblue}Ovis-VL-Embedding-2B & \color{revblue}Qwen3.5 & \color{revblue}35.51 & \color{revblue}60.69 & \color{revblue}44.34 & \color{revblue}4 & \color{revblue}32.01 & \color{revblue}57.24 & \color{revblue}40.62 & \color{revblue}6 \\
  & \color{revblue}EmbeddingGemma-2 & \color{revblue}Gemma4 & \color{revblue}26.54 & \color{revblue}50.20 & \color{revblue}34.60 & \color{revblue}10 & \color{revblue}21.34 & \color{revblue}43.73 & \color{revblue}29.01 & \color{revblue}17 \\
\cmidrule(lr){1-11}
\color{revblue}\multirow{6}{*}{$\geq$7B} & \color{revblue}Ops-MM-embedding-v1-7B & \color{revblue}Qwen2-VL & \color{revblue}24.09 & \color{revblue}45.04 & \color{revblue}31.16 & \color{revblue}16 & \color{revblue}25.01 & \color{revblue}46.17 & \color{revblue}32.09 & \color{revblue}15 \\
  & \color{revblue}Seed1.6-Embedding-1215 & \color{revblue}Seed1.6-Flash & \color{revblue}23.09 & \color{revblue}42.46 & \color{revblue}30.09 & \color{revblue}20 & \color{revblue}27.37 & \color{revblue}51.25 & \color{revblue}35.55 & \color{revblue}9 \\
  & \color{revblue}Qwen3-VL-Embedding-8B & \color{revblue}Qwen3-VL & \color{revblue}26.50 & \color{revblue}49.45 & \color{revblue}34.13 & \color{revblue}11 & \color{revblue}26.72 & \color{revblue}46.74 & \color{revblue}33.74 & \color{revblue}14 \\
  & \color{revblue}UEmbed-9B & \color{revblue}Qwen3.5 & \color{revblue}35.64 & \color{revblue}59.20 & \color{revblue}43.71 & \color{revblue}5 & \color{revblue}34.11 & \color{revblue}59.20 & \color{revblue}42.78 & \color{revblue}5 \\
  & \color{revblue}WeMM-Embedding-9B & \color{revblue}Qwen3.5 & \color{revblue}\textbf{45.39} & \color{revblue}\textbf{71.97} & \color{revblue}\textbf{54.76} & \color{revblue}\textbf{2} & \color{revblue}\textbf{38.78} & \color{revblue}\textbf{66.03} & \color{revblue}\textbf{48.01} & \color{revblue}\textbf{3} \\
  & \color{revblue}Ovis-VL-Embedding-9B & \color{revblue}Qwen3.5 & \color{revblue}41.71 & \color{revblue}67.95 & \color{revblue}50.30 & \color{revblue}3 & \color{revblue}38.52 & \color{revblue}64.58 & \color{revblue}47.51 & \color{revblue}\textbf{3} \\
\bottomrule
\end{tabular}
}
\label{tab_task1_main_results}
\end{table*}

\section{Experiments}
\label{sec_exp}

\subsection{Experimental Setup}

\paragraph{Models.} 
To establish a comprehensive benchmark, we evaluate a broad range of foundation models on MarsRetrieval.
We categorize the models into 3 groups.
(1)~\textbf{Dual-encoder vision-language models}: CLIP-DFN2B~\cite{fang2023data}, AIMv2-224~\cite{fini2025multimodal}, SigLIP~\cite{zhai2023sigmoid}, SigLIP2~\cite{tschannen2025siglip} and PE-Core~\cite{bolya2025perception}.
(2)~\textbf{MLLM-based embedding models}: \rev{jina-embeddings-v4~\cite{gunther2025jina}, Ops-MM-embedding~\cite{opensearch_mm_embedding_2b}, Qwen3-VL-Embedding~\cite{li2026qwen3}, UEmbed~\cite{song2026uembed}, WeMM-Embedding~\cite{zhou2026wemm}, Ovis-VL-Embedding~\cite{ovisembedding2026} and EmbeddingGemma-2~\cite{google2026embeddinggemma2}.}
(3) We also incorporate \textbf{Vision-only Foundation Models} like DINOv3~\cite{simeoni2025dinov3} and AIMv2-448~\cite{fini2025multimodal} for the image retrieval in Task 3.
Moreover, we include \textbf{MarScope}~\cite{wang2026natural}, a planetary science-specialized model fine-tuned from CLIP-DFN2B. 
For fair comparison, encoder-based models are selected at the ViT-L/14 or equivalent scale ($\sim$0.4B), \rev{while MLLM-based embedding models are evaluated at two scales ($\leq$3B and $\geq$7B) in Task~1 and at the largest available scale of each family in Task~2.}

\paragraph{Implementation Details.}
Across all three tasks, we adopt a unified embedding-based retrieval protocol. Images and texts are encoded independently using the corresponding encoders of each model.
Embedding models are evaluated using their official public checkpoints with default preprocessing and recommended input resolutions.
All embeddings are pre-computed offline and L2-normalized\rev{.}
During inference time, only the query embedding (text or image) is encoded and compared against the pre-indexed gallery using cosine similarity.

\subsection{Paired Image-Text Retrieval}

\paragraph{\rev{Scaling MLLM-based embedding models improves Martian image-text retrieval.}}
As shown in Table~\ref{tab_task1_main_results}, \rev{scaling MLLM-based embedding models from 2B to 7--9B consistently brings gains of 3.5--10.3 MRR points: Ops-MM improves from 28.40 to 32.09 on I$\to$T, while UEmbed improves from 32.50 to 42.78.}
\rev{Differences between model families are also large: WeMM-Embedding-2B outperforms Qwen3-VL-Embedding-8B by 11.30 MRR on T$\to$I.}
\rev{We further separate the effects of model size and input resolution in Appendix~\ref{appx_disc_scale}.}
\rev{Among general models without domain-specific finetuning, recent MLLM-based embedding models (e.g., WeMM-Embedding and Ovis-VL-Embedding) outperform PE-Core-L-14-336 in R@1 and MRR at both scales, and WeMM-Embedding-9B achieves the best R@1 and MRR in both directions (54.76 MRR on T$\to$I).}
Nevertheless, the absolute performance still remains moderate, indicating challenges in precisely ranking fine-grained, multiscale Martian image-text pairs.

\paragraph{\rev{Domain-specific post-training brings large gains.}}
\rev{MarScope achieves the best results among encoder-based models across both retrieval directions, leading the second-best PE-Core-L-14-336 by 10.19 MRR on T$\to$I and 7.80 on I$\to$T.}
It also improves substantially over its pre-trained version (CLIP-DFN2B), with gains of +29.52\% on R@10 (T$\to$I) and +22.57\% (I$\to$T), \rev{and attains the highest R@10 among all evaluated models.}
\rev{Fine-tuning other backbones yields similar gains, with each fine-tuned model favoring different tasks, as shown in Appendix~\ref{appx_disc_ft}.}

\subsection{Landform Retrieval}

\begin{table}[t]
\centering
\caption{Retrieval results of Landform Retrieval (task 2). \rev{MLLM-based models are reported at the largest available scale of each model family. Best results within each model group are in bold.}}
\resizebox{\linewidth}{!}{
\renewcommand\arraystretch{1.15}
\begin{tabular}{lccc}
\toprule
\textbf{Model} & \textbf{mAP} & \textbf{nDCG@10} & \textbf{Hits@10} \\
\midrule
\rowcolor{gray!15}\multicolumn{4}{c}{\textbf{\textit{Encoder-based Models}}} \\
\midrule
DFN2B-CLIP-ViT-L-14 & 9.48 & 12.39 & 47.92 \\
ViT-L-16-SigLIP-384 & 8.59 & 9.46 & 37.50 \\
aimv2-large-patch14-224 & 6.12 & 7.07 & 43.75 \\
ViT-L-16-SigLIP2-512 & 9.59 & 10.84 & 60.42 \\
PE-Core-L-14-336 & 20.93 & 25.80 & 72.92 \\
MarScope & \textbf{71.89} & \textbf{74.42} & \textbf{93.75} \\
\midrule
\rowcolor{gray!15}\multicolumn{4}{c}{\textbf{\textit{MLLM-based Models}}} \\
\midrule
\color{revblue}jina-embeddings-v4 & \color{revblue}6.57 & \color{revblue}4.67 & \color{revblue}31.25 \\
\color{revblue}Ops-MM-embedding-v1-7B & \color{revblue}10.76 & \color{revblue}10.27 & \color{revblue}39.58 \\
\color{revblue}Seed1.6-Embedding-1215 & \color{revblue}11.95 & \color{revblue}14.64 & \color{revblue}\textbf{58.33} \\
\color{revblue}Qwen3-VL-Embedding-8B & \color{revblue}10.09 & \color{revblue}12.53 & \color{revblue}47.92 \\
\color{revblue}UEmbed-9B & \color{revblue}11.38 & \color{revblue}14.27 & \color{revblue}54.17 \\
\color{revblue}WeMM-Embedding-9B & \color{revblue}13.83 & \color{revblue}17.60 & \color{revblue}52.08 \\
\color{revblue}Ovis-VL-Embedding-9B & \color{revblue}\textbf{16.61} & \color{revblue}\textbf{21.39} & \color{revblue}\textbf{58.33} \\
\color{revblue}EmbeddingGemma-2 & \color{revblue}7.94 & \color{revblue}8.92 & \color{revblue}39.58 \\
\bottomrule
\end{tabular}
}
\label{tab_task2_main}
\end{table}

\paragraph{Domain-specific fine-tuning is essential for Martian landform retrieval.}
Table~\ref{tab_task2_main} shows that Task~2 is substantially more challenging and domain-sensitive than generic image--text retrieval.
Among all models, MarScope achieves a dominant performance over all general pre-trained foundation models with $71.89\%$ for mAP and $93.75\%$ for Hits@10.
In contrast, even the strongest baseline (PE-Core-L-14-336) only reaches an mAP of $20.93\%$, which is nearly $50\%$ lower than that of MarScope.
This substantial gap highlights that general vision-language models lack the expertise of planetary science to distinguish subtle Martian geomorphic features.

\paragraph{MarScope's dominant performance is not driven by selection bias.}
To address potential bias for using MarScope during the initial data collection,  we partition all landform classes into two groups: those collected with MarScope (w/ MarScope, 38 classes) and those manually curated by planetary scientists from CTX and HiRISE imagery without MarScope (w/o MarScope, 10 classes), including \textit{Boulder Track, Fractured Mounds, Swiss Cheese}, etc. As shown in Table~\ref{tab_task2_marscope_bias}, MarScope's performance remains consistently dominant across both partitions\rev{, with 73.97 nDCG@10 on classes collected with MarScope and 76.43 on the others, and} no obvious performance gap between them.
These results demonstrate that the effectiveness of domain-specific models stems from geomorphic generalization rather than a \rev{``memorization''} of discovery-based patterns, thereby confirming the fairness of the MarsRetrieval evaluation.

\begin{table}[t]
\caption{Dataset bias analysis on Task~2. We report nDCG@10 for all 48 classes and subsets collected with and without MarScope. MarScope outperforms general models on both subsets.}
\centering
\resizebox{\linewidth}{!}{
\renewcommand\arraystretch{1.15}
\begin{tabular}{lccc}
\toprule
\color{revblue}\textbf{Model} & \color{revblue}\textbf{All} & \color{revblue}\textbf{w/o MarScope} & \color{revblue}\textbf{w/ MarScope} \\
\midrule
\color{revblue}SigLIP2 & \color{revblue}10.84 & \color{revblue}12.34 & \color{revblue}10.49 \\
\color{revblue}PE-Core & \color{revblue}25.80 & \color{revblue}24.86 & \color{revblue}26.02 \\
\color{revblue}MarScope & \color{revblue}\textbf{74.42} & \color{revblue}\textbf{76.43} & \color{revblue}\textbf{73.97} \\
\color{revblue}WeMM-9B & \color{revblue}17.60 & \color{revblue}10.51 & \color{revblue}19.23 \\
\color{revblue}Ovis-VL-9B & \color{revblue}21.39 & \color{revblue}12.04 & \color{revblue}23.61 \\
\bottomrule
\end{tabular}
}
\label{tab_task2_marscope_bias}
\end{table}

\subsection{Global Geo-Localization}

\begin{figure*}[t]
  \centering
  \includegraphics[width=\textwidth]{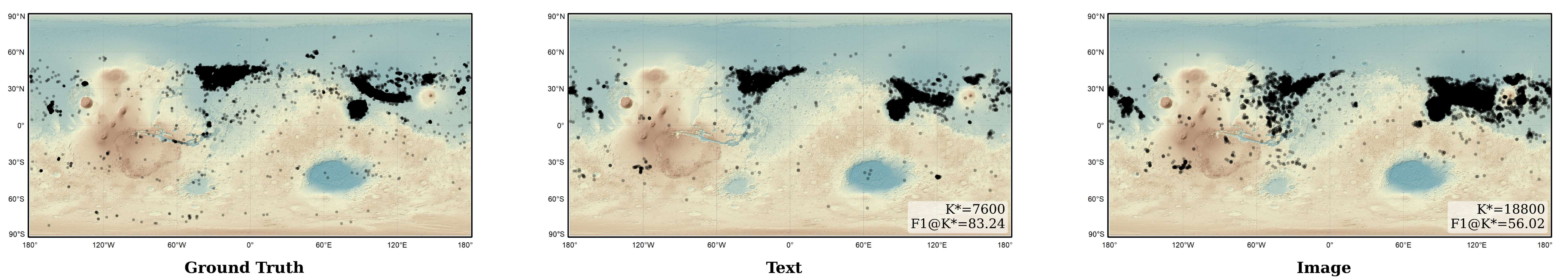}
  \caption{Global distribution of Pitted Cones in Task 3. The distribution retrieved by text-based query shows a more similar result to the ground truth catalog.}
  \label{fig_task3_distribution_pitted_cones}
\end{figure*}

\paragraph{Text-based retrieval can generalize to diverse Martian landforms after domain-specific fine-tuning.}
We provide the results for Task 3 (Global Geo-Localization) in Table~\ref{tab:task3_auprc_text_image}.
After domain-specific fine-tuning, \rev{MarScope achieves the strongest text-based retrieval performance on nearly all landform categories} (e.g., 14.88 on Alluvial Fans, 71.31 on Pitted Cones, and 65.38 on Yardangs), indicating robust alignment between geomorphology terminology and landform image patches.
We further visualize the predicted global distributions of Pitted Cones in Figure~\ref{fig_task3_distribution_pitted_cones}, where text-based predictions yield distributions substantially closer to the ground-truth catalog than image-based retrieval (see Figure~\ref{fig_task3_marscope_global_distribution} in Appendix~\ref{appx_exp} for all classes).
In contrast, general encoder-based VLMs and MLLM-based models reach only low AUPRC with text queries, which is consistent with the observation shown in Task~2. 
We provide more results on the optimal F1 score ($\text{F1}@{K^\star}$) in Appendix~\ref{appx_exp_task3} (See Table~\ref{tab_task3_f1_text_image} and Figure~\ref{fig_task3_marscope_f1_score}).
Overall, these results indicate that while general multimodal models possess broad knowledge, precise alignment between scientific terminology and Martian geomorphology requires explicit fine-tuning for the planetary domain.

\begin{table*}[t]
\caption{Retrieval results of Global Geo-Localization~(Task 3) on various foundation models. We report AUPRC for both text-based and image-based retrieval across five Martian landforms. \rev{Best results within the encoder-based and MLLM-based groups are in bold.}}
\centering
\resizebox{0.99\textwidth}{!}{
\renewcommand\arraystretch{1.15}
\begin{tabular}{lcc cc cc cc cc}
\toprule
\multirow{2}{*}{\textbf{Model}}
& \multicolumn{2}{c}{\textbf{Alluvial Fans}}
& \multicolumn{2}{c}{\textbf{Glacier-Like Forms}}
& \multicolumn{2}{c}{\textbf{Landslides}}
& \multicolumn{2}{c}{\textbf{Pitted Cones}}
& \multicolumn{2}{c}{\textbf{Yardangs}} \\
\cmidrule(lr){2-3}\cmidrule(lr){4-5}\cmidrule(lr){6-7}\cmidrule(lr){8-9}\cmidrule(lr){10-11}
& \textbf{Text} & \textbf{Image}
& \textbf{Text} & \textbf{Image}
& \textbf{Text} & \textbf{Image}
& \textbf{Text} & \textbf{Image}
& \textbf{Text} & \textbf{Image} \\
\midrule

\rowcolor{gray!25}\multicolumn{11}{c}{\textbf{\textit{Encoder-based VLMs}}} \\
\midrule
DFN2B-CLIP-ViT-L-14        &  1.47 &  2.32 &  1.96 & 17.90 & 29.27 &  4.76 &  7.29 & 16.71 & 0.24 & 28.86 \\
ViT-L-16-SigLIP-384        & 6.82 &  1.73 &  2.29 & 11.70 & 23.54 &  1.98 &  5.25 & 14.06 & 1.65 & 35.34 \\
aimv2-large-patch14-224    &  1.54 &  1.11 &  2.27 & 11.65 & 26.68 &  2.04 &  1.70 & 12.26 & 1.26 & 21.86 \\
ViT-L-16-SigLIP2-512       &  1.55 &  2.94 &  6.33 & 21.97 & 25.83 &  2.48 &  3.68 & 14.87 & 1.51 & 27.79 \\
PE-Core-L-14-336           &  3.82 &  2.23 & 6.66 & 14.24 & \textbf{31.14} &  6.79 & 16.86 & 19.91 & 2.63 & 34.20 \\
MarScope                   & \textbf{14.88} & \textbf{12.10} & \textbf{33.65} & \textbf{36.88} & 25.62 & \textbf{13.91} & \textbf{71.31} & \textbf{40.63} & \textbf{65.38} & \textbf{56.02} \\

\midrule
\rowcolor{gray!25}\multicolumn{11}{c}{\textbf{\textit{Vision-only Models}}} \\
\midrule
aimv2-large-patch14-448    &  -- &  2.38 &  -- & 18.03 &  -- & 8.30 &  -- & 14.23 &  -- & 19.47 \\
dinov3-vitl16              &  -- & 4.23 &  -- & 25.39 &  -- &  3.75 &  -- & 15.09 &  -- & 44.50 \\

\midrule
\rowcolor{gray!25}\multicolumn{11}{c}{\textbf{\textit{MLLM-based Models}}} \\
\midrule
jina-embeddings-v4 &  2.54 &  0.25 &  1.59 &  5.91 &  2.29 &  1.36 &  0.51 & 21.30 &  0.94 & 14.55 \\
Ops-MM-embedding-v1\rev{-2B} &  3.67 &  1.27 &  2.65 & 11.52 & 10.39 &  2.58 &  3.47 & 18.37 &  0.66 & 36.33 \\
Qwen3-VL-Embedding\rev{-2B} &  1.37 &  0.55 &  0.75 & 12.88 &  6.84 &  2.29 &  0.09 &  5.74 &  9.14 & 39.39 \\
\color{revblue}UEmbed\rev{-2B} & \color{revblue}2.53 & \color{revblue}\textbf{2.87} & \color{revblue}5.09 & \color{revblue}14.52 & \color{revblue}\textbf{31.42} & \color{revblue}\textbf{3.64} & \color{revblue}3.77 & \color{revblue}18.24 & \color{revblue}6.87 & \color{revblue}36.64 \\
\color{revblue}WeMM-Embedding\rev{-2B} & \color{revblue}2.98 & \color{revblue}1.05 & \color{revblue}8.67 & \color{revblue}13.19 & \color{revblue}11.24 & \color{revblue}1.25 & \color{revblue}\textbf{12.95} & \color{revblue}12.91 & \color{revblue}0.26 & \color{revblue}\textbf{42.17} \\
\color{revblue}Ovis-VL-Embedding\rev{-2B} & \color{revblue}1.60 & \color{revblue}1.69 & \color{revblue}\textbf{9.95} & \color{revblue}\textbf{16.66} & \color{revblue}7.02 & \color{revblue}2.30 & \color{revblue}12.42 & \color{revblue}17.31 & \color{revblue}2.28 & \color{revblue}36.35 \\
\color{revblue}EmbeddingGemma-2 & \color{revblue}\textbf{3.74} & \color{revblue}1.58 & \color{revblue}2.95 & \color{revblue}16.07 & \color{revblue}9.64 & \color{revblue}2.33 & \color{revblue}2.14 & \color{revblue}\textbf{23.86} & \color{revblue}\textbf{13.82} & \color{revblue}26.60 \\
\bottomrule
\end{tabular}
}
\label{tab:task3_auprc_text_image}
\end{table*}

\paragraph{Image-based retrieval remains competitive for \rev{general} vision models.}
To provide a comprehensive evaluation, we further conduct image-to-image retrieval using query instances curated by planetary scientists (details in Appendix~\ref{appx_task3_data}), ensuring that query images and database tiles are observed at the same spatial scale.
Interestingly, several general VLMs achieve competitive AUPRC under this setting and even outperform their text-based counterparts on specific classes (e.g., Pitted Cones, Yardangs).
Notably, the vision-only DINOv3-ViT-L/16 obtains the strongest overall image-based AUPRC among general models (25.39 on Glacier-Like Forms, 44.50 on Yardangs), confirming that general models already possess reasonable visual features for Martian discrimination but suffer from poor vision--language alignment in the planetary domain.
However, domain-specialized MarScope still performs substantially better with text queries than with image queries\rev{, reaching 71.31\% and 40.63\% on Pitted Cones, respectively. We attribute this to} the extreme intra-class morphological variation of Martian landforms---a small set of visual exemplars cannot fully characterize each category's diversity.
Overall, these findings suggest that improving planetary vision--language alignment is the key direction for scalable Mars geomorphology retrieval.

\section{Conclusion}
\rev{We introduce MarsRetrieval, the first comprehensive retrieval-centric benchmark for evaluating VLMs in planetary-scale geospatial discovery on Mars.
Its three complementary tasks, Paired Image-Text Retrieval, Landform Retrieval and Global Geo-Localization, bridge modern multimodal foundation models and language-driven workflows in planetary science.
Comprehensive benchmarking shows that MarsRetrieval remains highly challenging.
Recent MLLM-based embedding models perform well on paired image-text retrieval and benefit from scaling, but all general models fall far behind domain-specific models on landform retrieval and text-based geo-localization.
Domain-specific post-training brings large gains across all three tasks, suggesting that planetary vision-language alignment is the main bottleneck.
We hope MarsRetrieval serves as a standardized testbed for next-generation foundation models for Mars and broader deep-space exploration.}

\newpage

\section*{LLM Usage Statement}
\label{sec_llm_statement}
\rev{We used LLMs to polish the language of the manuscript, such as grammar, wording and sentence fluency. They were not used for ideation, experiment design, data analysis, writing code or generating results, and all scientific content was written and verified by the authors.
Separately, LLMs and MLLMs are part of the benchmark construction pipeline (\eg, data filtering and caption refinement in Task~1); this use is described in Section~\ref{task_1} and Appendix~\ref{appx_task1_data}.}

{
    \small
    \bibliographystyle{ieeenat_fullname}
    \bibliography{main}
}

\clearpage
\appendix
\etocdepthtag.toc{mtappendix}
\etocsettagdepth{mtbody}{none}
\etocsettagdepth{mtappendix}{subsubsection}
\etocsettocstyle{\noindent{\Large\bfseries Appendix}\par\vspace{4pt}\hrule\vspace{6pt}}{\vspace{4pt}\hrule\vspace{10pt}}
\twocolumn[{\tableofcontents}]

\section{Related Works}

\paragraph{Deep Learning in Planetary Science.} 
In recent years, the rapid progress of deep learning has substantially accelerated data analysis workflows in planetary science.
Enabled by advances in model architecture and the availability of large-scale orbital and rover-level imagery, several deep learning models have achieved competitive performance on Mars-related tasks including terrain classification~\cite{rothrock2016spoc, tiwari2024marsdeepnet, mengwall2023cloud}, feature analysis~\cite{zhao2025geological, zhou2025characteristics}, instance segmentation~\cite{xiong2023marsformer, li2025marsseg, delatte2019segmentation}, autonomous navigation~\cite{daftry2022mlnav, bechtold2023planetary}, global geology mapping~\cite{wang2026natural, fang2026domain, annex2025mars, pearson2024mapping}, etc.
Complementing these task-specific works, broader surveys argue that machine learning is becoming an essential tool for Martian geomorphology and exploration science~\cite{nagle2022squeezing}.

Despite recent advances, planetary machine learning still lacks standardized datasets and evaluation protocols that have catalyzed progress in other scientific domains. Existing benchmarks, such as Mars-Bench~\cite{purohit2025marsbench}, which consolidates 20 tasks spanning classification, segmentation, and object detection.
Alongside such broad benchmarks, specialized datasets have emerged to tackle specific geomorphic challenges, such as landslide segmentation~\cite{paheding2024marsls}, pitted cone segmentation~\cite{purohit2024conequest}.
However, these benchmarks primarily focus on closed-set supervised visual tasks, which are insufficient for assessing the cross-modal understanding required for modern language-driven planetary research.
MarsRetrieval addresses this gap by utilizing a retrieval-centric evaluation protocol. 
It provides curated datasets and standardized metrics for the evaluation of various vision-language models.
By doing so, MarsRetrieval provides a foundational resource for the emerging planetary ML community, facilitating reproducible research, fair model comparison, and the development of foundation model for deep-space exploration.

\paragraph{Vision-language Embedding Models.} 
Vision-Language models (VLMs) have revolutionized multimodal understanding by learning joint representations from vast image-text pairs.
Foundational architectures like CLIP~\cite{radford2021learning} and its variants, such as SigLIP~\cite{tschannen2025siglip} and PE-Core~\cite{bolya2025perception}, utilize dual-encoder and contrastive learning for cross-modal alignment by mapping images and text into a unified embedding space. 
The image and text representations can further be transferred to a wide range of downstream tasks (e.g., classification, retrieval, and generation).
Beyond dual encoders,  multimodal large language models (MLLMs)~\cite{liu2023visual, bai2025qwen2, chen2024internvl} couple a vision encoder with an autoregressive LLM to enable visual reasoning and instruction-following capabilities.
Recently, a growing line of work adapts MLLMs \rev{into embedding} models by extracting a fixed-dimensional representation from the LLM hidden states (e.g., pooling or the EOS/last-token state) and fine-tuning with retrieval-oriented contrastive learning, such as GME~\cite{zhang2024gme} and Qwen3-VL-Embedding~\cite{li2026qwen3}.
To systematically assess the utility of these diverse architectures in planetary science, we adopt a unified, retrieval-centric evaluation protocol.
We conduct an extensive evaluation of both dual-encoder VLMs and the MLLM-based embedding models across the 3 core tasks of MarsRetrieval to quantify their effectiveness for Martian geospatial discovery.

\begin{figure*}[t]
\centering
\begin{HeaderBox}{Prompt template for LLM-based filtering in Paired Image-Text Retrieval (Task 1)}
    You are a planetary science expert data classifier. \\
    Evaluate the following image descriptions. \\
    Determine if the image is related to Mars (the planet) or Martian context (surface, rovers, orbiters, geology). \\
    \\
    \textbf{Input Data:} \\
    $\langle$ \textit{image\_caption} $\rangle$ \\
    \\
    \textbf{Task:} \\
    - Answer ``Yes'' if the descriptions are related to Mars or Martian planetary science. \\
    - Answer ``No'' if they are NOT related to Mars or if the information is uncertain or ambiguous. \\
    \\
    Return ONLY one word: ``Yes'' or ``No''.
\end{HeaderBox}
\captionlistentry{}  
\label{prompt_llm_filtering}
\end{figure*}

\begin{figure*}[ht] %
\centering
\begin{HeaderBox}{Prompt template for MLLM-based filtering in Paired Image-Text Retrieval (Task 1)}
    \textbf{Instruction} \\
    You are an expert planetary science data classifier. Your task is to analyze the image and caption to determine if they are specifically related to Mars (the planet) or Martian context (surface, rovers, orbiters, geology). \\
    \\
    \textbf{Definitions} \\
    \textbf{YES (Mars or Martian Science):} \\
    - \textbf{Subjects:} Mars (the planet), Martian surface, Mars rovers (e.g., Perseverance), Mars orbiters, Martian geology. \\
    - \textbf{Surface Features:} Images related to Mars' surface features, such as craters, regolith, dunes, volcanoes, and valleys on Mars. \\
    - \textbf{Spacecraft:} Photos or data from Mars spacecraft (rovers, orbiters, landers) and scientific data from Mars missions. \\
    - \textbf{Scientific Data:} Spectra, geological maps, and any data related to Mars or its satellites. \\
    \\
    \textbf{NO (Not Mars or Martian Science):} \\
    - \textbf{Earth Context:} Any Earth-related content, such as Earth landscapes, cities, or forests. \\
    - \textbf{Astronomy or Astrophysics:} Deep space images, nebulae, galaxies, or objects unrelated to Mars. \\
    - \textbf{Artificial/Fake:} Sci-fi, CGI renders, video game screenshots, or concept art. \\
    - \textbf{Irrelevant:} Humans, selfies, museum exhibits, and Earth-based telescope equipment. \\
    \\
    \textbf{Constraint} \\
    - Focus strictly on Mars or Martian content. Exclude anything that is not directly related to Mars or its context. \\
    - If the image looks like a video game or art, answer ``no'' even if the caption mentions Mars. \\
    \\
    \textbf{Input Data} \\
    \textbf{Image:} $\langle$\textit{input\_image}$\rangle$ \\
    \textbf{Caption:} ``$\langle$\textit{image\_caption}$\rangle$'' \\
    \\
    Return ONLY one word: ``Yes'' or ``No''.
\end{HeaderBox}
\captionlistentry{}  
\label{prompt_mllm_filtering}
\end{figure*}

\begin{figure*}[t] %

\centering
\begin{HeaderBox}{Prompt template for MLLM-based caption refinement in Paired Image-Text Retrieval (Task 1)}
    You are a planetary science captioner for multimodal datasets. \\
    \\
    \textbf{Input Data} \\
    \textbf{IMAGE:} $\langle$\textit{input\_image}$\rangle$ \\
    \textbf{CAPTION:} ``$\langle$\textit{image\_caption}$\rangle$'' \\
    \\
    \textbf{Task:} \\
    Analyze the IMAGE and CAPTION to write ONE caption draft. \\
    \\
    \textbf{Hard Rules:} \\
    1) \textbf{Visual Grounding (PRIORITY):} \\
    - Describe visible attributes (shape, texture, brightness, etc.) derived from the IMAGE. \\
    - \textbf{Exception for Composition:} You MAY use composition terms ONLY if they are explicitly present in the caption and match the visual appearance. \\
    \\
    2) \textbf{Entity-First Structure:} \\
    - Start with the primary target / feature name from CAPTION. \\
    - \textbf{BANNED:} Do not start with ``The image shows'', ``A view of'', etc. \\
    \\
    3) \textbf{Anchor Logic:} \\
    - Use CAPTION for \textbf{Identity} (Subject Name, Location, Instrument). \\
    - Describe IMAGE with precise planetary geology terms. \\
    - If CAPTION describes features NOT visible in the crop, IGNORE them. \\
    \\
    4) \textbf{Noise Removal:} \\
    - Ignore IDs, coordinates, timestamps, and narrative fluff. \\
    \\
    \textbf{Style:} \\
    - Scientific, neutral, information-dense. \\
    \\
    \textbf{Return STRICT JSON only:} \\
    \{ ``refined\_caption'': ``$\langle$\textit{your\_caption}$\rangle$'' \}
\end{HeaderBox}
\captionlistentry{}  
\label{prompt_mllm_caption_fusion}
\end{figure*}

\section{Dataset Construction}

\subsection{Paired Image-Text Retrieval}
\label{appx_task1_data}

\subsubsection{Confidence Estimation}
\label{appx_task1_confidence}
For both the LLM-based domain relevance filtering and the VLM-based consistency filtering, the confidence score is derived from the model's output logits during a single-token inference step. 
we use a confidence score that represents the likelihood of a caption being related to Mars. 
Specifically, we extract the raw logits corresponding to the target tokens ``yes'' and ``no''. To obtain a confidence value, we apply a softmax function over these two logits to compute the normalized probability of the ``yes'' token:

$$ \text{Confidence} = \frac{\exp(L_{\text{yes}})}{\exp(L_{\text{yes}}) + \exp(L_{\text{no}})},$$

where $L_{\text{yes}}$ and $L_{\text{no}}$ represent the logit values for the respective tokens. This probability value serves as the final confidence score~\cite{luo2025your}. The prompt template for LLM and MLLM confidence are shown in Figure~\ref{prompt_llm_filtering} and Figure~\ref{prompt_mllm_filtering}.

\subsubsection{Data Filtering}
\label{appx_task1_filtering}
As we mentioned in Section~\ref{task1_data_operation}, we implemented a data filtering strategy based on three metrics to ensure the relevance and quality of the image-text pairs in our dataset construction of task 1. 
Specifically, we apply multi-modal filtering using the following thresholds:
 the CLIP score for text-image semantic alignment with a threshold of $\tau_{clip} = 0.32$, the LLM relevance confidence for Mars-specific grounding with a threshold of $\tau_{llm} = 0.82$, and the MLLM consistency score for visual support of the claimed Martian features with a threshold of $\tau_{mllm} = 0.82$.
A candidate pair is retained only when all three scores exceed the respective thresholds.

\begin{revblock}
\subsubsection{Data Statistics at Each Stage}
\label{appx_task1_stats}
Table~\ref{tab_task1_stages} reports the number of image-text pairs retained at each construction stage of Task~1.
The candidate pool is retrieved from two web-scale corpora that are not curated for Mars, so it contains duplicates, non-Mars content and mismatched image-text pairs, and each stage targets a distinct issue.
Caption deduplication removes exact or near-duplicate captions.
The score-based filtering removes pairs with weak visual grounding, low Mars relevance or inconsistent image-text content: by their primary rejection reason, 33,321 pairs show insufficient image-text alignment or visual grounding, 10,122 lack Mars relevance, 1,140 contain ineligible content and 958 have overly generic captions.
The K-center sampling then removes redundant pairs rather than low-quality ones, maximizing coverage in the text and image embedding spaces, and the final expert validation verifies the scientific validity and pair correctness.

\begin{table}[h]
\centering
\caption{Number of image-text pairs at each stage of Task~1.}
\label{tab_task1_stages}
\resizebox{0.8\linewidth}{!}{%
\begin{tabular}{lcc}
\toprule
\textbf{Stage} & \textbf{Input} & \textbf{Retained} \\
\midrule
Caption deduplication & 86,182 & 61,868 \\
Score-based filtering & 61,868 & 16,327 \\
Text K-center sampling & 16,327 & 5,000 \\
Image K-center sampling & 5,000 & 2,500 \\
Two-stage expert validation & 2,500 & 2,287 \\
\bottomrule
\end{tabular}}
\end{table}

\subsubsection{Ablation on Individual Score Filters}
\label{appx_task1_score_ablation}
To test whether any single score is sufficient for filtering, we construct fixed 5,000-pair subsets with no score, with each score alone and with all three scores jointly.
We use Seed2.0-Lite to judge the sampled pairs under three random sampling seeds (42, 43 and 44), and a pair is accepted only if it is Mars-relevant, image-text consistent, eligible scientific imagery and sufficiently distinctive.
As shown in Table~\ref{tab_task1_score_ablation}, the joint three-score filter achieves the highest overall acceptance rate (46.47\%), indicating that the three scores provide complementary signals.

\begin{table*}[t]
\centering
\caption{Acceptance rates (\%) of 5,000-pair subsets filtered with no score, each score alone and all three scores, judged by Seed2.0-Lite. The overall acceptance is pooled over three sampling seeds, with the Wilson 95\% confidence interval.}
\label{tab_task1_score_ablation}
\begin{tabular}{lccccc}
\toprule
\textbf{Condition} & \textbf{Mars Relevant} & \textbf{Image-Text Consistent} & \textbf{Eligible} & \textbf{Distinctive} & \textbf{Overall Accept (95\% CI)} \\
\midrule
No score & 45.67 & 82.00 & 42.67 & 69.33 & 25.87 [23.71, 28.14] \\
CLIP only & 45.60 & 82.33 & 42.33 & 72.47 & 33.87 [31.78, 36.09] \\
LLM only & 48.67 & 78.13 & 45.87 & 69.47 & 35.67 [33.52, 37.94] \\
MLLM only & 52.27 & 76.07 & 53.67 & 72.13 & 39.53 [37.28, 41.89] \\
All three & \textbf{61.33} & 79.93 & \textbf{59.20} & \textbf{74.73} & \textbf{46.47} [44.07, 48.93] \\
\bottomrule
\end{tabular}
\end{table*}

\subsubsection{Coverage of Omitted Candidates}
\label{appx_task1_coverage}
To quantify how much information is lost through data reduction, we compute, for each omitted candidate, its cosine similarity to the nearest sample among the final 2,287 pairs, using Qwen3-Embedding for text and DINOv3 for images.
The mean nearest-neighbor similarity is 0.9051 for text and 0.8411 for images, indicating that the final set retains broad semantic and visual coverage of the candidate pool.
While this metric cannot capture all forms of information loss, it suggests that most omitted content remains represented in the final set.

\subsubsection{Sensitivity to Filtering Thresholds}
\label{appx_task1_threshold}
To test the sensitivity to the filtering thresholds, we vary only the CLIP, LLM and MLLM thresholds of the score-based filtering and apply the same diversity sampling to each candidate pool.
We use Seed2.0-Lite to judge pair quality and report the Spearman correlation between the T$\to$I MRR ranking of four models (SigLIP2, PE-Core, MarScope and Qwen3-VL) under each setting and under the default setting (Table~\ref{tab_task1_threshold}).
Stricter thresholds improve the audited quality but reduce coverage, while the model ranking remains stable.
We further perform 10,000 paired query-bootstrap replicates for T$\to$I retrieval on each fixed 3,500-pair threshold-specific subset (Table~\ref{tab_task1_threshold_ci}); these subsets are constructed from the initial candidate pool rather than from the final benchmark.
The model ranking is identical under all three thresholds, with all adjacent paired 95\% confidence intervals excluding zero.
We therefore adopt the default thresholds as a balance between quality and coverage.

\begin{table*}[t]
\centering
\caption{Sensitivity to the score-based filtering thresholds. Audit acceptance is judged by Seed2.0-Lite, and rank correlation is the Spearman correlation of the T$\to$I MRR ranking of four models with the default setting.}
\label{tab_task1_threshold}
\begin{tabular}{lcccc}
\toprule
\textbf{Setting} & \textbf{CLIP / LLM / MLLM} & \textbf{Filtered Pool} & \textbf{Audit Accept, \% (95\% CI)} & \textbf{Rank Corr.} \\
\midrule
Lenient & 0.30 / 0.78 / 0.78 & 39,061 & 42.4 [38.4, 46.6] & 0.8 \\
Default & 0.32 / 0.82 / 0.82 & 16,327 & 46.0 [41.9, 50.3] & 1.0 \\
Strict & 0.34 / 0.86 / 0.86 & 7,957 & 61.0 [56.8, 65.4] & 1.0 \\
\bottomrule
\end{tabular}
\end{table*}

\begin{table*}[t]
\centering
\caption{T$\to$I MRR (\%) with 95\% confidence intervals on fixed 3,500-pair subsets constructed under each threshold setting (10,000 paired query-bootstrap replicates).}
\label{tab_task1_threshold_ci}
\begin{tabular}{lccc}
\toprule
\textbf{Setting} & \textbf{Qwen3-VL} & \textbf{SigLIP2} & \textbf{PE-Core} \\
\midrule
Lenient & 40.06 [38.57, 41.55] & 45.63 [44.22, 47.05] & 51.43 [50.08, 52.80] \\
Default & 36.98 [35.50, 38.44] & 41.66 [40.23, 43.09] & 47.83 [46.44, 49.24] \\
Strict & 25.53 [24.11, 26.96] & 27.91 [26.49, 29.33] & 31.95 [30.53, 33.37] \\
\bottomrule
\end{tabular}
\end{table*}
\end{revblock}

\subsection{Landform Retrieval}
{\setlength{\parindent}{0pt}%
\subsubsection{Prompt Ensemble}
\label{appx_task2_prompt_ensemble}
In Landform Retrieval (Task 2), we employ a prompt ensembling strategy to generate a robust prototype query embedding for each geomorphic subclass by averaging the output embeddings from the model's text encoder across multiple linguistic variations. To rigorously evaluate the sensitivity of our retrieval-centric protocol to these variations, we conducted an ablation study using ten distinct candidate templates, as detailed in Table~\ref{tab_task2_prompt_ablation}. These templates are designed to capture diverse scientific and observational perspectives, specifically: ``a photo of {}, a type of martian terrain'', ``a satellite photo of {}'', ``a high-resolution remote sensing image of {} on Mars'', ``a detailed satellite view of {} geomorphic feature from Mars'', ``an aerial perspective of {} morphological structure on Mars'', ``high-definition remote sensing photo of {} on the red planet'', ``a CTX or HiRISE image showing {} terrain on Mars'', ``an orbital image of {} landform on the Martian surface'', ``Martian {} as captured in space-based imagery'', and ``planetary-scale capture of {} in Martian geology''. Our experiments demonstrate that while increasing the number of templates can yield performance gains for domain-specialized models like MarScope, the incremental benefit for most general foundation models tends to saturate at three templates. Based on these findings, we adopted an ensemble of the first 3 templates as the default setting throughout the paper.
}

\subsection{Global Geo-Localization}
\label{appx_task3_data}

\begin{figure}[t]
  \centering
  \includegraphics[width=0.99\columnwidth]{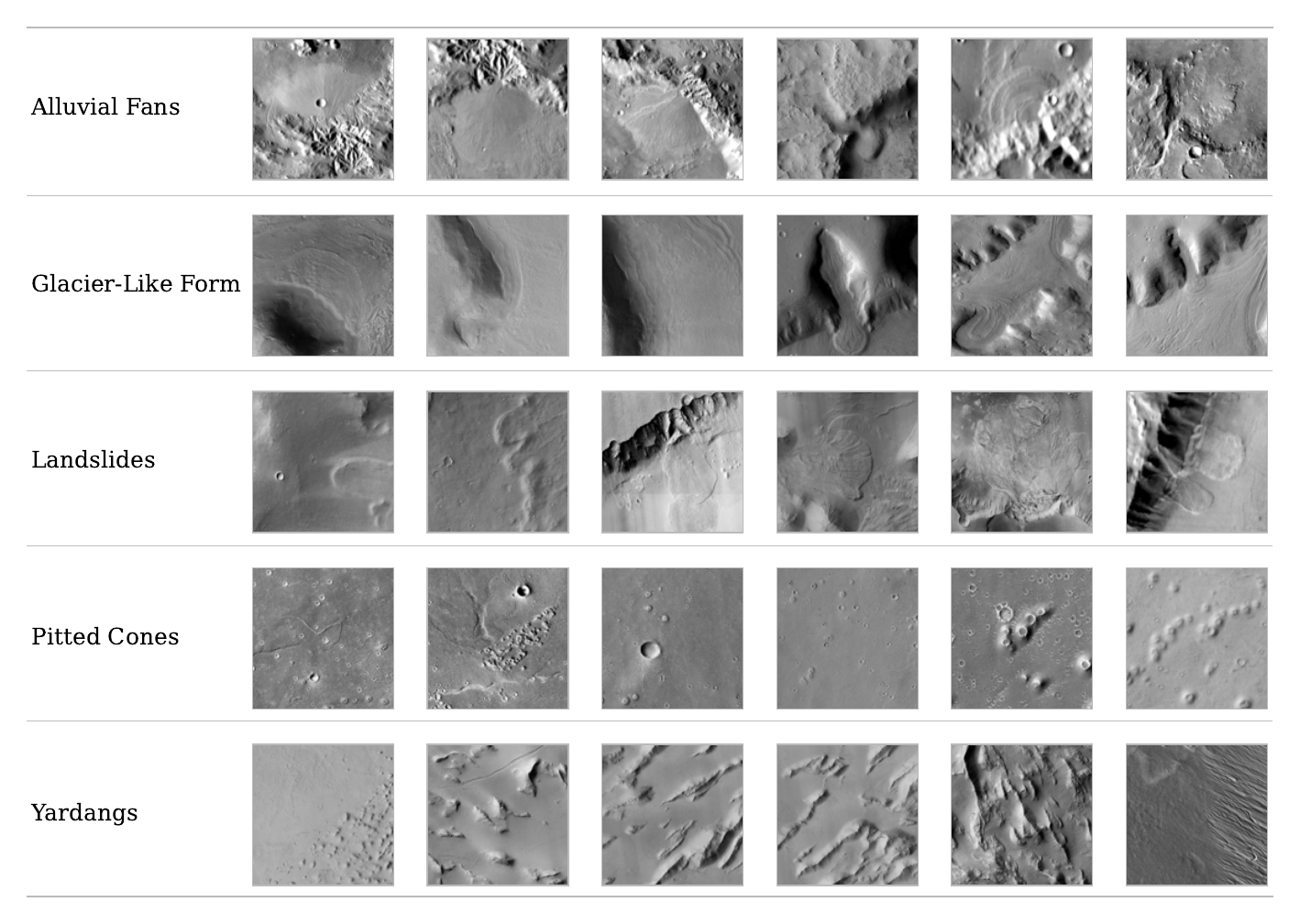}
  \caption{Image-based query examples for Task 3 (Global Geo-Localization)}
  \label{fig_task3_distribution}
\end{figure}

{\setlength{\parindent}{0pt}%
\subsubsection{Image-based Retrieval}
\label{appx_task3_image_query}
To provide a comprehensive evaluation of model performance in Global Geo-Localization, we utilize both text-based and image-based queries to estimate the planetary-scale distribution of Martian landforms. 
For each target landform category, we select a small set of query images (Alluvial Fans: 8, Glacier-Like Form: 8, Landslides: 10, Pitted cones: 8, Yardangs: 18), which is manually curated by planetary scientists.
For visualization, we show 6 random query examples per class in Figure~\ref{fig_task3_distribution}.
}

\begin{revblock}
\subsubsection{Relation Between MarScope and the Benchmark}
\label{appx_task3_marscope}
According to~\cite{wang2026natural}, MarScope is post-trained on image-text pairs from official archives (NASA Photojournal and HiRISE), figures and captions from planetary-science publications, and public media (NASA Flickr and SciTechDaily).
These training sources do not include the five Task~3 catalogues or their coordinate annotations, and the Task~3 gallery is constructed separately by tiling the global CTX mosaic at a fixed spatial scale rather than by reusing MarScope's training images.
\end{revblock}

\section{Evaluation Metrics}
\label{appx_metrics}

\subsection{Paired Image-Text Retrieval}
\label{appx_task1_metrics}
To systematically evaluate the cross-modal alignment capabilities of Vision-Language Models (VLMs) in the planetary science domain, we formulate Task 1 as a bidirectional retrieval problem. 
Given a set of $N$ paired Martian image-text instances $\mathcal{D} = \{(i_k, t_k)\}_{k=1}^N$, for each query in one modality (e.g., a text description $t_j$), the model is required to rank all $N$ candidates in the opposite modality (e.g., images $\{i_1, \dots, i_N\}$) based on their cosine similarity. 
The retrieval performance is evaluated on several widely-used metrics that capture both the accuracy and quality of ranking:

{\setlength{\parindent}{0pt}%
\paragraph{\bfseries\upshape Recall@K (R@K)}
This metric measures the proportion of queries for which the ground-truth match appears within the top-K ranked candidates, which can be formally defined as:

$$R@K = \frac{1}{N} \sum_{j=1}^{N} \mathbb{I}(r_j \le K),$$

}

where $\mathbb{I}(\cdot)$ is the indicator function. We report \textbf{R@1} and  \textbf{R@10} to assess high-precision and top-tier retrieval performance.

{\setlength{\parindent}{0pt}%
\paragraph{\bfseries\upshape Mean Reciprocal Rank (MRR)}
Mean Reciprocal Rank (MRR) rewards the models for ranking the correct match as high as possible and is defined as:
$$MRR = \frac{1}{N} \sum_{j=1}^{N} \frac{1}{r_j}.$$
}

{\setlength{\parindent}{0pt}%
\paragraph{\bfseries\upshape Median Rank (MedR)}
To evaluate the overall stability of the model's representation space and mitigate the influence of extreme outliers, we report the Median Rank (MedR):
$$\text{MedR} = \text{median}(\{r_1, r_2, \dots, r_N\}).$$
}

A lower MedR indicates a more robust and consistent alignment across the entire paired vision-language retrieval dataset.

\subsection{Landform Retrieval}
\label{appx_task2_metrics}
We evaluate the task of landform retrieval in a text-to-image multi-positive setting, where each query is encoded from a Martian landform and needs to retrieve multiple relevant images. 
Let $\{x_1, x_2, \dots, x_N\}$ be the ranked list of database images for query $q$, and $\mathrm{rel}_i \in \{0, 1\}$ be the binary relevance indicator where $\mathrm{rel}_i = 1$ if $x_i \in \mathcal{G}_q$. 
We employ the following metrics to conduct evaluation under the long-tailed distribution of the 48 categories:

{\setlength{\parindent}{0pt}%
\paragraph{\bfseries\upshape Macro Mean Average Precision (mAP)}
We evaluate retrieval performance using mean Average Precision (mAP) computed as a uniform (macro) average over $C$ text queries $\{q_c\}_{c=1}^{C}$ (one per landform category). 
For a query $q_c$, given the ranked list of database images $\{x_i\}_{i=1}^{N}$ and relevance indicators
$\mathrm{rel}_i=\mathbb{1}[x_i\in\mathcal{G}_{q_c}]$, the Average Precision (AP) is:
}

\[
\mathrm{AP}(q_c)=\frac{1}{|\mathcal{G}_{q_c}|}\sum_{i=1}^{N} P@i \cdot \mathrm{rel}_i,
\quad
P@i=\frac{\sum_{j=1}^{i}\mathrm{rel}_j}{i}.
\]

\[
\mathrm{mAP}=\frac{1}{C}\sum_{c=1}^{C}\mathrm{AP}(q_c).
\]

{\setlength{\parindent}{0pt}%
\paragraph{\bfseries\upshape nDCG@10}
To evaluate whether the most representative instances are prioritized in the top-tier results, we report the Normalized Discounted Cumulative Gain at cutoff 10. For the query $q_c$ associated with category $c$:

\[
\text{DCG@10}(q_c) = \sum_{i=1}^{10} \frac{\mathrm{rel}_i}{\log_2(i+1)},
\]
\[
\text{nDCG@10} = \frac{1}{C} \sum_{c=1}^{C} \frac{\text{DCG@10}(q_c)}{\text{IDCG@10}(q_c)},
\]
}
where $\text{IDCG@10}(q_c)$ is the ideal DCG obtained by ranking all available positives for $q_c$ at the top of the list (up to the cutoff 10) and applying the same DCG@10 definition.

{\setlength{\parindent}{0pt}%
\paragraph{\bfseries\upshape Hits@10}
This measures the model's ability to find at least one correct instance within the top 10 results:

$$\text{Hits@10} = \frac{1}{C} \sum_{q=1}^{C} \mathbb{I}\left(\sum_{i=1}^{10} \mathrm{rel}_i > 0\right).$$
}

\begin{figure*}[t]
  \includegraphics[width=0.99\textwidth]{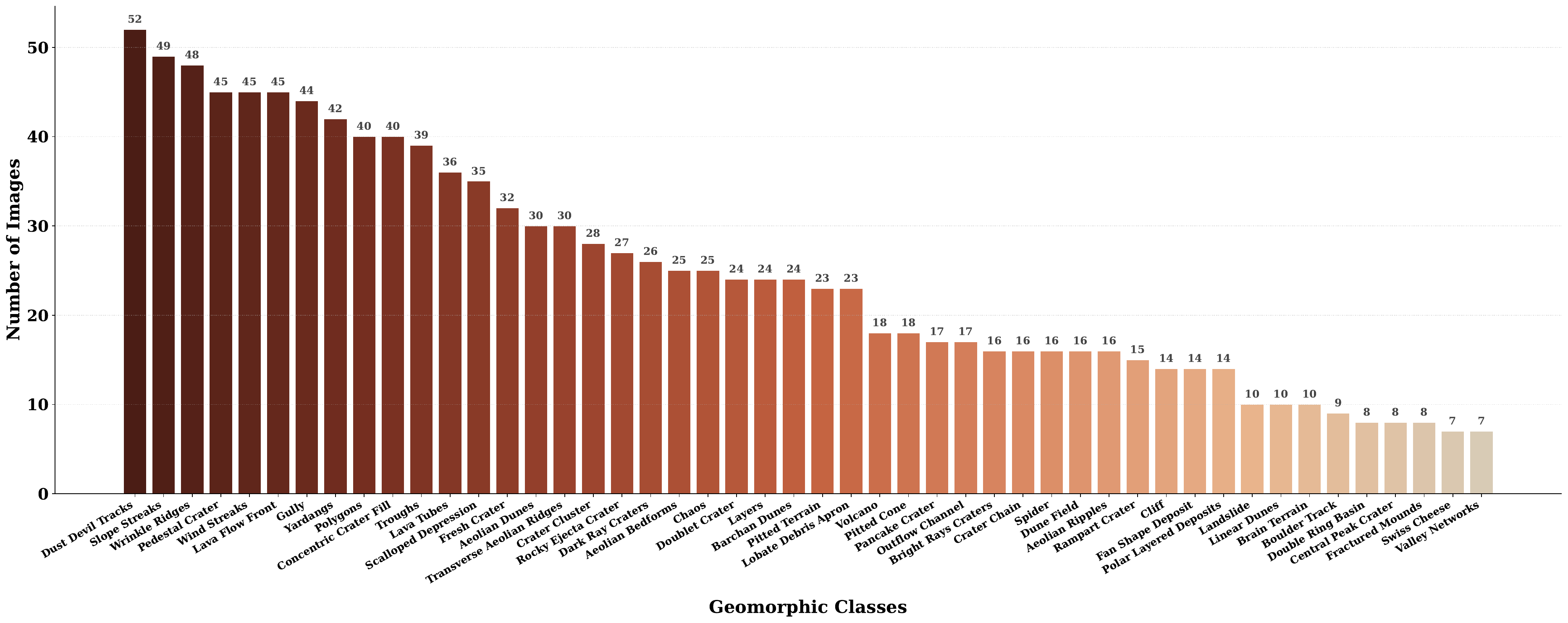}
  \caption{Sample distribution of the 48 long-tailed geomorphic categories utilized in Landform Retrieval (Task 2).}
  \label{fig_task2_detailed_distribution}
\end{figure*}

\subsection{Global Geo-Localization}
\label{appx_task3_metrics}
In Task 3, we evaluate a model's ability to estimate the planetary-scale spatial distribution of Martian landforms. Given the extreme sparsity of positive samples within the global CTX mosaic ($\sim 1.4$ million tiles), we formulate this as a proximity-based point-set matching problem in a projected coordinate space.

To ensure computational efficiency across the global dataset, we map all geographic coordinates $(\text{lon}, \text{lat})$ onto a global equirectangular (Plate Carrée) mosaic with a fixed resolution. A retrieved image patch is considered a true positive if its projected center coordinate $\mathbf{p}'$ falls within a pixel-wise Euclidean distance $r_{px}$ of a projected ground-truth entry $\mathbf{g}'$:
$$\|\mathbf{p}' - \mathbf{g}'\|_2 \le r_{px}.$$
The pixel-wise tolerance $r_{px}$ is derived from a spatial angular radius $r$ (set to $0.5^{\circ}$) as: $r_{px} = r \cdot (W / 360^{\circ})$. 

For a retrieval depth $K$, let $\mathcal{P}'_K = \{\mathbf{p}'_1, \mathbf{p}'_2, \dots, \mathbf{p}'_K\}$ be the projected coordinates of the top-$K$ retrieved tiles, and $\mathcal{G}' = \{\mathbf{g}'_1, \mathbf{g}'_2, \dots, \mathbf{g}'_M\}$ be the $M$ projected reference points from the catalogue. We define the foundational metrics as follows:

{\setlength{\parindent}{0pt}%
\paragraph{\bfseries\upshape Precision@K}
This metric measures retrieval fidelity, representing the fraction of predicted tiles spatially validated by the catalogue in the projected space:
$$\text{Precision}(K) = \frac{1}{K} \sum_{i=1}^{K} \mathbb{I}\left( \min_{\mathbf{g}' \in \mathcal{G}'} \| \mathbf{p}'_i - \mathbf{g}' \|_2 \le r_{px} \right).$$
}

{\setlength{\parindent}{0pt}%
\paragraph{\bfseries\upshape Recall@K}
This metric measures mapping completeness, representing the fraction of catalogue entries successfully recovered within the tolerance radius:
$$\text{Recall}(K) = \frac{1}{M} \sum_{j=1}^{M} \mathbb{I}\left( \min_{\mathbf{p}' \in \mathcal{P}'_K} \| \mathbf{p}' - \mathbf{g}'_j \|_2 \le r_{px} \right).$$
}

Based on the Precision and Recall defined above, we report the following cumulative metrics to characterize the global localization performance:

{\setlength{\parindent}{0pt}%
\paragraph{\bfseries\upshape Area Under the Precision-Recall Curve (AUPRC)}
AUPRC assesses the trade-off between retrieval precision and recall across all sensitivity levels (retrieval depths). We define it as:
$$
\text{AUPRC} = \int_{0}^{1} \text{Precision}(\rho)\, d\rho,\quad \rho=\text{Recall}.
$$
In practice, the PR curve is obtained by sweeping the retrieval depth $K$, and the integral is approximated using trapezoidal integration over the sampled points $\{(\text{Recall}(K), \text{Precision}(K))\}_K$.
}

{\setlength{\parindent}{0pt}%
\paragraph{\bfseries\upshape $\text{F1}@K^\star$}
To quantify the optimal balance between mapping fidelity and completeness, we report the maximum F1-score achievable on the PR curve. The F1-score at depth $K$ is defined as:
$$\text{F1}(K) = 2 \cdot \frac{\text{Precision}(K) \cdot \text{Recall}(K)}{\text{Precision}(K) + \text{Recall}(K)}.$$
}
We specifically report the peak performance $\text{F1}@K^\star = \max_K \text{F1}(K)$, where $K^\star$ represents the optimal retrieval depth for the specific model.

\section{Experimental Results}
\label{appx_exp}

\subsection{Paired Image-Text Retrieval}
\label{appx_exp_task1}

\subsubsection{Effect of Caption Refinement}
\label{appx_exp_task1_caption}
\begin{figure}[t]
    \centering
    \includegraphics[width=\linewidth]{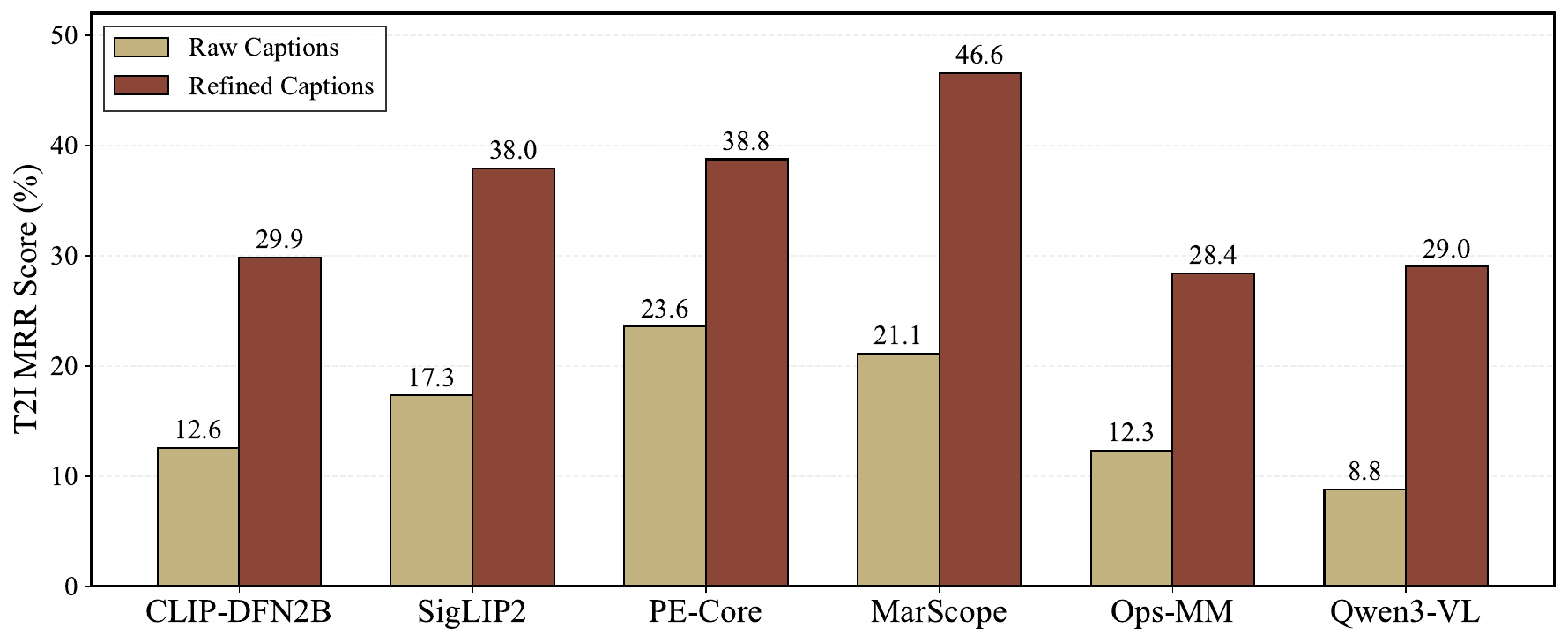}
    \caption{Ablation on caption refinement in Paired Image--Text Retrieval (task~1). The results on MRR indicate high-quality captioning leads to stronger vision--language alignment.}
    \label{fig_task1_ablation_caption}
\end{figure}

\paragraph{Caption refinement consistently improves retrieval quality.} 
As illustrated in Figure~\ref{fig_task1_ablation_caption}, all evaluated models yield consistent MRR gains when applying the refined captions (See Section~\ref{task1_data_operation}), with MarScope and SigLIP2 benefiting the most (e.g., +25.4 and +20.6 MRR, respectively).
The results show that the refined captions can successfully eliminate image-text irrelevant noise and provide more precise and discriminative descriptions of Martian features.
We present the results with the raw caption in Table~\ref{tab_task1_raw_results}.
Overall, the results demonstrate that the quality of captions is crucial for vision-language alignment in planetary science tasks.

\begin{table*}[t]
\caption{Performance comparison of different models on Paired Image-Text Retrieval (Task 1) with raw (unrefined) captions. Notably, performance across all models is significantly lower compared to the results with refined captions (see Table~\ref{tab_task1_main_results}). Bold indicates the best result within each model group.}
\centering
\resizebox{0.95\textwidth}{!}{
\renewcommand\arraystretch{1.15}
\begin{tabular}{lcc cccc cccc}
\toprule
\multirow{2}{*}{\textbf{Model}} & \multirow{2}{*}{\textbf{Backbone}} & \multirow{2}{*}{\textbf{Size}} & \multicolumn{4}{c}{\textbf{T$\to$I}} & \multicolumn{4}{c}{\textbf{I $\to$ T}} \\
\cmidrule(lr){4-7}\cmidrule(lr){8-11}
 &   &   & \textbf{R@1} & \textbf{R@10} & \textbf{MRR} & \textbf{MedR} & \textbf{R@1} & \textbf{R@10} & \textbf{MRR} & \textbf{MedR} \\
\midrule
\rowcolor{gray!25}\multicolumn{11}{c}{\textbf{\textit{Encoder-based Models}}} \\
\midrule
\color{revblue}DFN2B-CLIP-ViT-L-14 & \color{revblue}ViT-L/14 & \color{revblue}0.4B & \color{revblue}8.26 & \color{revblue}22.47 & \color{revblue}12.92 & \color{revblue}116 & \color{revblue}8.05 & \color{revblue}20.73 & \color{revblue}12.58 & \color{revblue}130 \\
\color{revblue}ViT-L-16-SigLIP-384 & \color{revblue}ViT-L/16 & \color{revblue}0.4B & \color{revblue}11.28 & \color{revblue}22.43 & \color{revblue}15.27 & \color{revblue}167 & \color{revblue}9.01 & \color{revblue}20.24 & \color{revblue}13.02 & \color{revblue}176 \\
\color{revblue}aimv2-large-patch14-224 & \color{revblue}ViT-L/14 & \color{revblue}0.4B & \color{revblue}4.68 & \color{revblue}14.43 & \color{revblue}8.11 & \color{revblue}233 & \color{revblue}4.77 & \color{revblue}14.12 & \color{revblue}8.09 & \color{revblue}241 \\
\color{revblue}ViT-L-16-SigLIP2-512 & \color{revblue}ViT-L/16 & \color{revblue}0.4B & \color{revblue}13.16 & \color{revblue}27.72 & \color{revblue}18.21 & \color{revblue}89 & \color{revblue}12.42 & \color{revblue}26.98 & \color{revblue}17.34 & \color{revblue}87 \\
\color{revblue}PE-Core-L-14-336 & \color{revblue}ViT-L/14 & \color{revblue}0.4B & \color{revblue}\textbf{17.01} & \color{revblue}\textbf{38.35} & \color{revblue}\textbf{24.42} & \color{revblue}\textbf{24} & \color{revblue}\textbf{16.44} & \color{revblue}\textbf{37.04} & \color{revblue}\textbf{23.57} & \color{revblue}\textbf{27} \\
\color{revblue}MarScope (DFN2B) & \color{revblue}ViT-L/14 & \color{revblue}0.4B & \color{revblue}13.73 & \color{revblue}37.69 & \color{revblue}21.77 & \color{revblue}29 & \color{revblue}14.25 & \color{revblue}34.67 & \color{revblue}21.15 & \color{revblue}39 \\
\midrule
\rowcolor{gray!25}\multicolumn{11}{c}{\textbf{\textit{MLLM-based Models}}} \\
\midrule
\color{revblue}gme & \color{revblue}Qwen2-VL & \color{revblue}2B & \color{revblue}9.44 & \color{revblue}19.59 & \color{revblue}13.07 & \color{revblue}220 & \color{revblue}7.52 & \color{revblue}16.40 & \color{revblue}10.76 & \color{revblue}\textbf{286} \\
\color{revblue}jina-embeddings-v4 & \color{revblue}Qwen2.5-VL & \color{revblue}3B & \color{revblue}8.61 & \color{revblue}15.52 & \color{revblue}11.36 & \color{revblue}336 & \color{revblue}5.95 & \color{revblue}13.12 & \color{revblue}8.65 & \color{revblue}377 \\
\color{revblue}Ops-MM-embedding-v1 & \color{revblue}Qwen2-VL & \color{revblue}2B & \color{revblue}\textbf{11.98} & \color{revblue}\textbf{21.86} & \color{revblue}\textbf{15.77} & \color{revblue}\textbf{180} & \color{revblue}\textbf{9.14} & \color{revblue}\textbf{18.23} & \color{revblue}\textbf{12.33} & \color{revblue}287 \\
\color{revblue}Qwen3-VL-Embedding & \color{revblue}Qwen3-VL & \color{revblue}2B & \color{revblue}9.75 & \color{revblue}18.10 & \color{revblue}12.87 & \color{revblue}272 & \color{revblue}6.03 & \color{revblue}14.08 & \color{revblue}8.82 & \color{revblue}372 \\
\bottomrule
\end{tabular}
}
\label{tab_task1_raw_results}
\end{table*}

\begin{revblock}
\subsubsection{Model-Family Bias in Data Curation}
\label{appx_exp_task1_family}
Since Qwen models are used for filtering and diversity sampling while Qwen-family models are also evaluated, we test whether this curation favors Qwen-family models by replacing Qwen with the Seed2.0 family at each stage.
\textbf{Score-based filtering.}
Seed2.0-Lite independently judges 1,000 pairs retained by Qwen and 1,000 pairs rejected by Qwen: it rejects 164 of the retained pairs but accepts only 12 of the rejected pairs, indicating that Qwen-based filtering is more permissive than Seed rather than biased toward excluding valid samples.
\textbf{Diversity sampling.}
Rebuilding a 2,500-pair candidate pool with the Seed embedding model doubao-embedding-vision-251215 instead of Qwen3-Embedding changes 48.2\% of the selected samples, yet the ranking of six representative models is preserved.
Overall, we find no evidence that Qwen-based curation favors Qwen-family models.

\end{revblock}

\subsection{Landform Retrieval}
\label{appx_exp_task2}

\begin{table}[t]
\centering
\caption{Ablation on different prompt templates in Landform Retrieval (task 2). We use 3 templates throughout the paper.}
\resizebox{\linewidth}{!}{
\renewcommand\arraystretch{1.15}
\begin{tabular}{lccccc}
\toprule
\textbf{Model} & \textbf{1} & \textbf{3} & \textbf{5} & \textbf{7} & \textbf{10} \\
\midrule
MarScope              & 68.74 & 71.89 & 71.87 & \textbf{72.44} & 72.32 \\
PE-Core-L-14-336      & 20.67 & \textbf{20.93} & 20.06 & 19.68 & 19.84 \\
Qwen3-VL-Embedding    & 8.43  & \textbf{8.69}  & 7.63  & 7.73  & 7.51 \\
Ops-MM-embedding-v1   & 9.52  & \textbf{9.62}  & 7.28  & 8.35  & 8.50 \\
\bottomrule
\end{tabular}
}
\label{tab_task2_prompt_ablation}
\end{table}

\subsubsection{Effect of Prompt Ensemble}
\label{appx_exp_task2_prompt}
\paragraph{Prompt ensemble provides consistent performance gains.} 
To verify the effectiveness of the prompt ensemble mentioned in Section~\ref{task2_setup}, we provide an ablation study in Table~\ref{tab_task2_prompt_ablation}.
For MarScope, ensembling 7 templates yields the highest macro-mAP of 72.44\%, a gain of +3.70\% over the single-prompt baseline.
Moreover, the incremental gains for other general models tend to saturate at 3 templates.
To maintain an optimal balance between retrieval accuracy and simplicity, we adopt an ensemble of three prompts as the default setting throughout our main evaluation.

\begin{revblock}
\subsubsection{Earth-Shared and Planetary-Specific Terms}
\label{appx_exp_task2_term}
To test whether the Task~2 gains of MarScope come mainly from learning planetary terms, we split the 48 landform categories into 22 categories whose names are also used for Earth landforms or have a direct Earth equivalent (\eg, Barchan Dunes, Gully, Yardangs) and 26 categories mainly used in planetary science (\eg, Rampart Crater, Swiss Cheese, Lobate Debris Apron).
As shown in Table~\ref{tab_task2_term}, fine-tuning yields similar nDCG@10 gains for both groups (+61.39 and +62.59), indicating that the improvement is not limited to unfamiliar planetary vocabulary.

\begin{table}[h]
\centering
\caption{Task~2 nDCG@10 of DFN2B-CLIP-ViT-L-14 and MarScope on Earth-shared and planetary-specific categories.}
\label{tab_task2_term}
\resizebox{0.9\linewidth}{!}{%
\begin{tabular}{lcccc}
\toprule
\textbf{Category Group} & \textbf{\#Classes} & \textbf{DFN2B-CLIP} & \textbf{MarScope} & \textbf{Gain} \\
\midrule
Earth-shared & 22 & 15.85 & 77.25 & +61.39 \\
Planetary-specific & 26 & 9.47 & 72.05 & +62.59 \\
\bottomrule
\end{tabular}}
\end{table}
\end{revblock}

\subsection{Global Geo-Localization}
\label{appx_exp_task3}

\subsubsection{Optimal F1 Score}
\label{appx_exp_task3_f1}
{\setlength{\parindent}{0pt}%
\paragraph{\bfseries\upshape Optimal F1 further emphasizes the importance of domain-specific post-training.}
Table~\ref{tab_task3_f1_text_image} provides a detailed comparison of the optimal F1-score ($F1@K^*$) across various models for Task 3 (Global Geo-Localization). 
The results highlight the dominance of the domain-specialized model MarScope, \rev{which outperforms general foundation models on nearly all landform categories.}
This gap underscores that general multimodal representations lack the specialized planetary knowledge required to establish precise links between scientific terminology and subtle geomorphic structures.
In summary, without explicit planetary-domain fine-tuning, generic representations are insufficient for precise planetary-scale discovery.
}

\subsubsection{Visualization of Global Distributions}
\label{appx_exp_task3_vis}
\begin{figure*}[t]
  \includegraphics[width=0.99\textwidth]{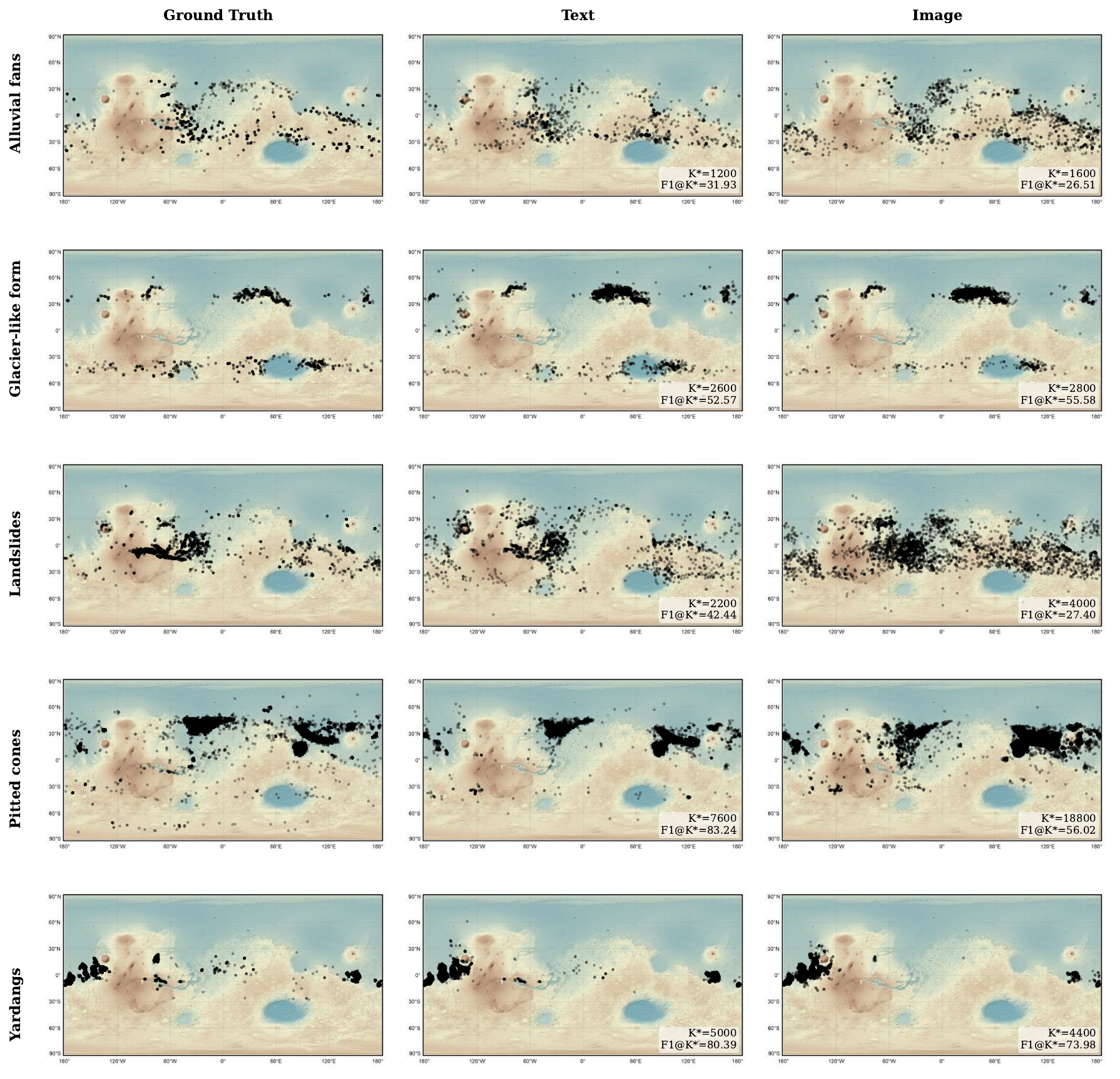}
  \caption{The visualization of planetary-scale geomorphic distributions generated by MarScope (Task 3). MarScope's text-based retrieval produces distributions that are more consistent with global catalogues compared to image-based retrieval.}
  \label{fig_task3_marscope_global_distribution}
\end{figure*}

{\setlength{\parindent}{0pt}%
\paragraph{\bfseries\upshape Text queries produce distributions more consistent with global catalogues.}
To qualitatively assess retrieval effectiveness, we visualize the global geomorphic distributions generated by MarScope in Figure~\ref{fig_task3_marscope_global_distribution}.
\rev{Compared with the human-curated ground-truth catalogs, the global distributions predicted by MarScope closely follow the catalogued occurrences.}
For categories such as Pitted Cones and Glacier-like forms, text queries accurately capture the latitudinal distribution of these landforms.
While image-based retrieval successfully identifies major clusters, it is more prone to false positives from geomorphically similar but unrelated terrains.
These results support the potential of language-driven retrieval for scalable planetary mapping.
}

\begin{figure}[t]
\centering

\begin{subfigure}[b]{0.49\linewidth}
\includegraphics[width=\textwidth]{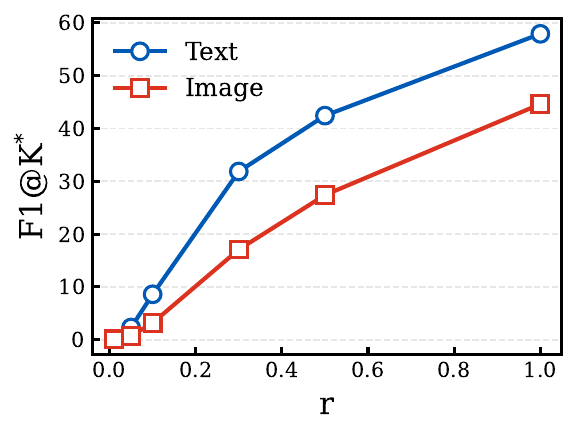}
\caption{Landslides}
\end{subfigure}
\hfill
\begin{subfigure}[b]{0.49\linewidth}
\includegraphics[width=\textwidth]{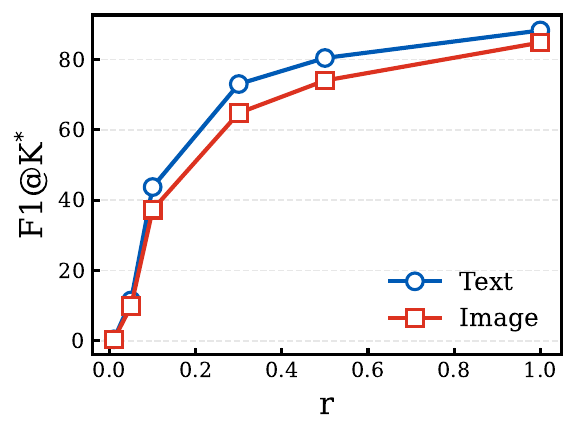}
\caption{Yardangs}
\end{subfigure}

\caption{Ablation on the spatial tolerance radius $r$ in Global Geo-Localization~(task 3). We use $r{=}0.5$ throughout the paper.}
\label{fig_task3_ablation_r}
\end{figure}
\subsubsection{Sensitivity to Retrieval Depth}
\label{appx_exp_task3_depth}
\begin{figure*}[t]
  \includegraphics[width=0.99\textwidth]{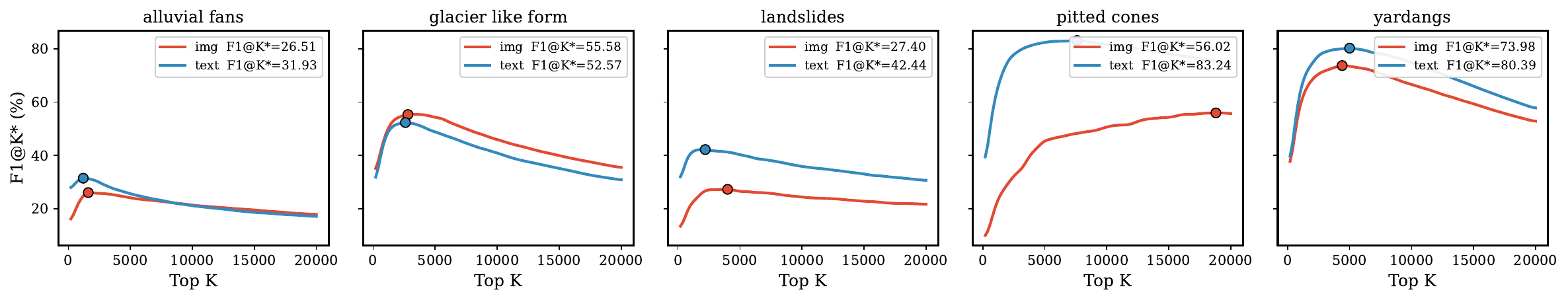}
  \caption{Sensitivity of ($F1@K$) to retrieval depth $K$ in task 3. Text-based queries exhibit better peak performance and stability.}
  \label{fig_task3_marscope_f1_score}
\end{figure*}

\paragraph{\bfseries\upshape Text-based retrieval shows stronger peak performance and better stability.}
Figure~\ref{fig_task3_marscope_f1_score} illustrates the sensitivity of the F1-score to the retrieval depth $K$ in Task 3.
The results shows that text queries typically achieve higher peaks and e and superior stability across different values of $K$.
This reflects the inherent intra-class visual diversity of Martian landforms, where a limited set of image exemplars cannot fully represent the morphological evolution of a category across the entire planet.

\begin{table*}[t]
\caption{Performance comparison of Global Geo-Localization (Task 3) measured by the optimal F1-score ($F1@K^*$). \rev{The domain-specialized MarScope outperforms general foundation models on most landform categories.} This gap highlights the necessity of explicit fine-tuning with planetary knowledge for precise large-scale scientific discovery. \rev{Best results within the encoder-based and MLLM-based groups are in bold.}}
\centering
\resizebox{\textwidth}{!}{
\renewcommand\arraystretch{1.15}
\begin{tabular}{lcc cc cc cc cc}
\toprule
\multirow{2}{*}{\textbf{Model}}
& \multicolumn{2}{c}{\textbf{Alluvial Fans}}
& \multicolumn{2}{c}{\textbf{Glacier-Like Forms}}
& \multicolumn{2}{c}{\textbf{Landslides}}
& \multicolumn{2}{c}{\textbf{Pitted Cones}}
& \multicolumn{2}{c}{\textbf{Yardangs}} \\
\cmidrule(lr){2-3}\cmidrule(lr){4-5}\cmidrule(lr){6-7}\cmidrule(lr){8-9}\cmidrule(lr){10-11}
& \textbf{Text} & \textbf{Image}
& \textbf{Text} & \textbf{Image}
& \textbf{Text} & \textbf{Image}
& \textbf{Text} & \textbf{Image}
& \textbf{Text} & \textbf{Image} \\
\midrule

\rowcolor{gray!25}\multicolumn{11}{c}{\textbf{\textit{Encoder-based Models}}} \\
\midrule
DFN2B-CLIP-ViT-L-14        &  6.19 &  7.94 &  7.16 & 34.98 & 47.90 & 13.69 & 19.38 & 36.52 &  2.92 & 47.11 \\
ViT-L-16-SigLIP-384        & 16.26 &  6.71 &  7.69 & 24.89 & 38.73 &  8.95 & 17.12 & 29.13 &  7.97 & 51.75 \\
aimv2-large-patch14-224    &  6.83 &  4.67 &  8.02 & 24.76 & 42.63 &  8.84 &  9.51 & 28.80 &  6.41 & 38.01 \\
ViT-L-16-SigLIP2-512       &  7.13 &  9.38 & 15.32 & 39.40 & 42.96 &  9.82 & 15.47 & 32.32 &  8.01 & 46.21 \\
PE-Core-L-14-336           & 13.82 &  7.47 & 15.02 & 29.54 & \textbf{48.82} & 16.36 & 34.69 & 36.93 &  8.94 & 51.47 \\
MarScope                   & \textbf{31.93} & \textbf{26.51} & \textbf{52.57} & \textbf{55.58} & 42.44 & \textbf{27.40} & \textbf{83.24} & \textbf{56.02} & \textbf{80.39} & \textbf{73.98} \\

\midrule
\rowcolor{gray!25}\multicolumn{11}{c}{\textbf{\textit{Vision-only Models}}} \\
\midrule
aimv2-large-patch14-448    &  -- &  7.42 &  -- & 33.96 &  -- & 18.86 &  -- & 33.28 &  -- & 35.04 \\
dinov3-vitl16              &  -- & 12.27 &  -- & 43.06 &  -- & 12.33 &  -- & 32.50 &  -- & 62.95 \\

\midrule
\rowcolor{gray!25}\multicolumn{11}{c}{\textbf{\textit{MLLM-based Models}}} \\
\midrule
jina-embeddings-v4 &  7.34 &  2.22 &  5.87 & 15.32 &  8.80 &  6.00 &  4.98 & \textbf{39.99} &  5.45 & 27.73 \\
Ops-MM-embedding-v1\rev{-2B} &  9.26 &  5.15 &  8.49 & 25.26 & 22.56 &  9.34 & 13.76 & 36.55 &  3.77 & 55.12 \\
Qwen3-VL-Embedding\rev{-2B} &  5.84 &  3.29 &  3.77 & 26.41 & 17.52 &  8.83 &  3.00 & 18.13 & 21.40 & 57.49 \\
\color{revblue}UEmbed\rev{-2B} & \color{revblue}8.72 & \color{revblue}\textbf{8.22} & \color{revblue}14.26 & \color{revblue}30.01 & \color{revblue}\textbf{48.55} & \color{revblue}\textbf{11.65} & \color{revblue}14.59 & \color{revblue}35.52 & \color{revblue}19.11 & \color{revblue}54.00 \\
\color{revblue}WeMM-Embedding\rev{-2B} & \color{revblue}8.89 & \color{revblue}5.23 & \color{revblue}19.10 & \color{revblue}27.89 & \color{revblue}23.12 & \color{revblue}6.38 & \color{revblue}28.67 & \color{revblue}26.90 & \color{revblue}3.25 & \color{revblue}\textbf{58.64} \\
\color{revblue}Ovis-VL-Embedding\rev{-2B} & \color{revblue}6.10 & \color{revblue}6.49 & \color{revblue}\textbf{21.03} & \color{revblue}\textbf{32.28} & \color{revblue}17.60 & \color{revblue}8.95 & \color{revblue}\textbf{29.04} & \color{revblue}33.35 & \color{revblue}10.51 & \color{revblue}53.27 \\
\color{revblue}EmbeddingGemma-2 & \color{revblue}\textbf{12.86} & \color{revblue}6.26 & \color{revblue}9.48 & \color{revblue}31.51 & \color{revblue}22.65 & \color{revblue}9.22 & \color{revblue}11.07 & \color{revblue}39.95 & \color{revblue}\textbf{29.37} & \color{revblue}46.05 \\
\bottomrule
\end{tabular}
}
\label{tab_task3_f1_text_image}
\end{table*}

\subsubsection{Spatial Tolerance Radius}
\label{appx_exp_task3_radius}
\paragraph{The spatial tolerance radius is essential for reasonable geo-localization evaluation.}
Following the evaluation protocol mentioned in Section~\ref{task3_setup}, we conduct an ablation study on the spatial tolerance radius $r$ to verify its necessity.
We present the $F1@K^*$ scores for Landslides and Yardangs across a range of $r$ values in Figure~\ref{fig_task3_ablation_r}.
When $r$ is set to an extremely small value (e.g., $r = 0.01^{\circ}$), the $F1$ scores remain near zero (e.g., $0.13\%$ for Landslides), making it difficult to distinguish model performance or establish scientific significance.
As $r$ increases, the metric becomes substantially more informative and stable.
Hence, to balance the trade-off between localization performance and benchmark fairness, we adopt $r=0.5^{\circ}$ as the default setting in this paper.

\begin{revblock}
\section{Discussion}
\label{appx_discussion}

\subsection{\rev{Fine-Tuning Across Backbones}}
\label{appx_disc_ft}
To test whether the gains of MarScope are specific to its CLIP backbone, we \rev{fine-tune ViT-L-16-SigLIP2-512 and Qwen3-VL-Embedding-2B on Mars image-text pairs from} the training sources of MarScope~\cite{wang2026natural} and evaluate them on all three tasks (Table~\ref{tab_disc_ft}).

Fine-tuning improves every backbone on every task; for instance, Task~2 mAP rises from below 10 to between 48.00 and 71.89.
The fine-tuned models nevertheless show distinct task profiles: Qwen3-VL-Embedding achieves the best Task~1 results (55.15 MRR on T$\to$I and 48.12 on I$\to$T), whereas MarScope and SigLIP2 perform better on Tasks~2 and~3.
Thus, MarsRetrieval distinguishes among fine-tuned models rather than only separating fine-tuned from zero-shot models.

\begin{table*}[t]
\centering
\caption{Zero-shot and domain-specific fine-tuned results of three backbones. Task~1 reports MRR with Seed2.0 captions, Task~2 reports mAP, and Task~3 reports AUPRC averaged over the five landform categories. Best fine-tuned results are in bold.}
\label{tab_disc_ft}
\resizebox{0.8\textwidth}{!}{%
\begin{tabular}{llccccc}
\toprule
& & \multicolumn{2}{c}{\textbf{Task 1 (MRR)}} & \textbf{Task 2} & \multicolumn{2}{c}{\textbf{Task 3 (AUPRC)}} \\
\cmidrule(lr){3-4}\cmidrule(lr){5-5}\cmidrule(lr){6-7}
\textbf{Backbone} & \textbf{Setting} & \textbf{T$\to$I} & \textbf{I$\to$T} & \textbf{mAP} & \textbf{Text} & \textbf{Image} \\
\midrule
\multirow{2}{*}{DFN2B-CLIP-ViT-L-14} & Zero-shot & 25.67 & 29.87 & 9.48 & 8.05 & 14.11 \\
 & Fine-tuned (MarScope) & 47.70 & 46.57 & \textbf{71.89} & \textbf{42.17} & 31.91 \\
\midrule
\multirow{2}{*}{ViT-L-16-SigLIP2-512} & Zero-shot & 34.06 & 37.95 & 9.59 & 7.78 & 14.01 \\
 & Fine-tuned & 44.37 & 42.83 & 68.13 & 41.85 & \textbf{34.15} \\
\midrule
\multirow{2}{*}{Qwen3-VL-Embedding-2B} & Zero-shot & 29.85 & 29.03 & 8.68 & 3.64 & 12.17 \\
 & Fine-tuned & \textbf{55.15} & \textbf{48.12} & 48.00 & 28.21 & 25.56 \\
\bottomrule
\end{tabular}}
\end{table*}

\subsection{Statistical Uncertainty of Rankings}
\label{appx_disc_ci}
We estimate 95\% confidence intervals \rev{of the Task~1 and Task~2 scores} with a paired bootstrap of 10,000 resamples: for Task~1 we resample the image-text pairs with replacement and use the same resampled indices for all models, and for Task~2 we resample the landform categories.
Table~\ref{tab_disc_ci} reports the Task~1 MRR intervals of all models.

\rev{The Task~1 intervals are narrow, with half-widths of 1.2--1.8 MRR, and the Task~2 mAP gap between MarScope and PE-Core-L-14-336 is 50.96 [41.87, 59.91].}

\begin{table*}[t]
\centering
\caption{Task~1 MRR with paired bootstrap 95\% confidence intervals (10,000 resamples of the 2,287 image-text pairs, Seed2.0 captions).}
\label{tab_disc_ci}
\resizebox{0.8\linewidth}{!}{%
\begin{tabular}{lcccc}
\toprule
& \multicolumn{2}{c}{\textbf{T$\to$I}} & \multicolumn{2}{c}{\textbf{I$\to$T}} \\
\cmidrule(lr){2-3}\cmidrule(lr){4-5}
\textbf{Model} & \textbf{MRR} & \textbf{95\% CI} & \textbf{MRR} & \textbf{95\% CI} \\
\midrule
\multicolumn{5}{l}{\textit{Encoder-based Models}} \\
DFN2B-CLIP-ViT-L-14 & 25.67 & [24.19, 27.15] & 29.87 & [28.29, 31.46] \\
ViT-L-16-SigLIP-384 & 29.20 & [27.58, 30.81] & 29.78 & [28.20, 31.41] \\
aimv2-large-patch14-224 & 16.83 & [15.60, 18.08] & 19.15 & [17.86, 20.49] \\
ViT-L-16-SigLIP2-512 & 34.06 & [32.40, 35.77] & 37.95 & [36.25, 39.67] \\
PE-Core-L-14-336 & 37.51 & [35.79, 39.15] & 38.76 & [37.10, 40.49] \\
MarScope & 47.70 & [46.01, 49.39] & 46.57 & [44.87, 48.25] \\
\midrule
\multicolumn{5}{l}{\textit{MLLM-based Models ($\leq$3B)}} \\
jina-embeddings-v4 & 22.98 & [21.44, 24.51] & 23.08 & [21.59, 24.58] \\
Ops-MM-embedding-v1\rev{-2B} & 27.68 & [26.13, 29.30] & 28.40 & [26.77, 30.03] \\
Qwen3-VL-Embedding\rev{-2B} & 29.85 & [28.24, 31.47] & 29.03 & [27.38, 30.69] \\
UEmbed\rev{-2B} & 34.90 & [33.17, 36.60] & 32.50 & [30.84, 34.22] \\
WeMM-Embedding\rev{-2B} & 45.43 & [43.66, 47.23] & 42.72 & [40.96, 44.51] \\
Ovis-VL-Embedding\rev{-2B} & 44.34 & [42.56, 46.12] & 40.62 & [38.91, 42.40] \\
EmbeddingGemma-2 & 34.60 & [32.91, 36.30] & 29.01 & [27.43, 30.61] \\
\midrule
\multicolumn{5}{l}{\textit{MLLM-based Models ($\geq$7B)}} \\
Ops-MM-embedding-v1\rev{-7B} & 31.16 & [29.50, 32.86] & 32.09 & [30.43, 33.81] \\
Seed1.6-Embedding-1215 & 30.09 & [28.42, 31.76] & 35.55 & [33.86, 37.27] \\
Qwen3-VL-Embedding\rev{-8B} & 34.13 & [32.43, 35.83] & 33.74 & [32.03, 35.50] \\
UEmbed\rev{-9B} & 43.71 & [41.91, 45.49] & 42.78 & [41.05, 44.60] \\
WeMM-Embedding\rev{-9B} & 54.76 & [52.98, 56.56] & 48.01 & [46.22, 49.82] \\
Ovis-VL-Embedding\rev{-9B} & 50.30 & [48.52, 52.12] & 47.51 & [45.74, 49.32] \\
\midrule
\multicolumn{5}{l}{\textit{Fine-tuned Backbones (Appendix~\ref{appx_disc_ft})}} \\
ViT-L-16-SigLIP2-512 (fine-tuned) & 44.37 & [42.68, 46.11] & 42.82 & [41.10, 44.57] \\
Qwen3-VL-Embedding-2B (fine-tuned) & 55.15 & [53.47, 56.89] & 48.12 & [46.44, 49.80] \\
\bottomrule
\end{tabular}}
\end{table*}

\subsection{Training Randomness of MarScope}
\label{appx_disc_seeds}
To estimate the variance caused by training randomness, we retrain the MarScope recipe~\cite{wang2026natural} with seeds 7, 8 and 9 under the same data, hyperparameters and checkpoint-selection rule.
These replicas use an expanded version of the training corpus, so they measure the stability of the training recipe rather than the uncertainty of the released checkpoint.

As shown in Table~\ref{tab_disc_seeds}, the results are stable on Tasks~2 and~3 (standard deviation below 2.5 points), whereas Task~1 varies more across seeds (4.82 MRR on T$\to$I).
\rev{The seed variance is small compared with the gains of the MarScope recipe on Tasks~2 and~3 (\eg, a 50.96 mAP lead over PE-Core-L-14-336 on Task~2).}

\begin{table*}[t]
\centering
\caption{MarScope trained with three random seeds. Task~1 reports MRR with Seed2.0 captions, Task~2 reports mAP and nDCG@10, and Task~3 reports AUPRC averaged over the five landform categories. The last row reports mean $\pm$ sample standard deviation.}
\label{tab_disc_seeds}
\resizebox{0.8\textwidth}{!}{%
\begin{tabular}{lcccccc}
\toprule
& \multicolumn{2}{c}{\textbf{Task 1 (MRR)}} & \multicolumn{2}{c}{\textbf{Task 2}} & \multicolumn{2}{c}{\textbf{Task 3 (AUPRC)}} \\
\cmidrule(lr){2-3}\cmidrule(lr){4-5}\cmidrule(lr){6-7}
\textbf{Seed} & \textbf{T$\to$I} & \textbf{I$\to$T} & \textbf{mAP} & \textbf{nDCG@10} & \textbf{Text} & \textbf{Image} \\
\midrule
7 & 35.91 & 34.18 & 67.52 & 70.57 & 43.66 & 30.90 \\
8 & 44.76 & 37.09 & 70.52 & 74.45 & 42.42 & 29.32 \\
9 & 37.04 & 31.37 & 70.89 & 74.31 & 40.02 & 32.09 \\
\midrule
Mean $\pm$ std & 39.24 $\pm$ 4.82 & 34.21 $\pm$ 2.86 & 69.64 $\pm$ 1.85 & 73.11 $\pm$ 2.20 & 42.04 $\pm$ 1.85 & 30.77 $\pm$ 1.39 \\
\bottomrule
\end{tabular}}
\end{table*}

\begin{table*}[t]
\centering
\caption{Earth-observation CLIP models and their pre-trained backbones. Task~1 reports MRR with Seed2.0 captions, Task~2 reports nDCG@10, and Task~3 reports AUPRC averaged over the five landform categories.}
\label{tab_disc_eo}
\resizebox{0.85\textwidth}{!}{%
\begin{tabular}{llccccc}
\toprule
& & \multicolumn{2}{c}{\textbf{Task 1 (MRR)}} & \textbf{Task 2} & \multicolumn{2}{c}{\textbf{Task 3 (AUPRC)}} \\
\cmidrule(lr){3-4}\cmidrule(lr){5-5}\cmidrule(lr){6-7}
\textbf{Model} & \textbf{Adapted On} & \textbf{T$\to$I} & \textbf{I$\to$T} & \textbf{nDCG@10} & \textbf{Text} & \textbf{Image} \\
\midrule
OpenAI CLIP-ViT-L-14 & Generic & 15.32 & 20.19 & 9.98 & 0.82 & 13.35 \\
\quad + RemoteCLIP & Earth observation & 8.55 & 17.51 & 3.42 & 0.57 & 12.37 \\
\quad + SkyCLIP & Earth observation & 16.52 & 20.92 & 11.90 & 1.95 & 11.47 \\
\midrule
LongCLIP-ViT-L-14 & Generic & 21.93 & 21.88 & 9.14 & 5.15 & 16.69 \\
\quad + DGTRS-CLIP & Earth observation & 15.44 & 19.82 & 10.48 & 3.75 & 15.83 \\
\midrule
DFN2B-CLIP-ViT-L-14 & Generic & 25.67 & 29.87 & 12.39 & 8.05 & 14.11 \\
\quad + MarScope & Mars & \textbf{47.70} & \textbf{46.57} & \textbf{74.42} & \textbf{42.17} & \textbf{31.91} \\
\bottomrule
\end{tabular}}
\end{table*}

\subsection{Earth-Observation Baselines}
\label{appx_disc_eo}
To test whether adaptation to Earth remote-sensing data transfers to Mars, we evaluate three ViT-L/14 CLIP models fine-tuned on Earth-observation image-text data: RemoteCLIP~\cite{liu2024remoteclip}, SkyCLIP~\cite{wang2024skyscript} and DGTRS-CLIP~\cite{chen2025dgtrs}, together with their pre-trained backbones.

As shown in Table~\ref{tab_disc_eo}, Earth-observation adaptation brings no consistent improvement over the corresponding backbone: RemoteCLIP and DGTRS-CLIP are worse than their backbones on most metrics, SkyCLIP improves by at most 1.92 points, and all three models remain far below MarScope.
We further search the released training captions of these models for the three Task~3 landforms that also occur on Earth: out of 0.66M to 1.76M captions per corpus, yardangs are never mentioned and alluvial fans and landslides appear in at most 88 captions.
These results suggest that the gains of MarScope come from Mars-specific data rather than from general remote-sensing adaptation.

\subsection{Model Size and Input Resolution}
\label{appx_disc_scale}
Larger encoders are often evaluated at higher resolutions, which confounds the effect of model size.
We therefore evaluate PE-Core-L-14 (native resolution 336) and PE-Core-G-14 (native resolution 448) at the same three input resolutions (Table~\ref{tab_disc_scale}).

At matched resolution, G/14 outperforms L/14 by 1.29--4.64 MRR on Task~1 and by 3.80--6.00 nDCG@10 on Task~2 at 336 and 448, but degrades sharply at 224, far below its native resolution.
Thus, a larger model helps when evaluated at a compatible resolution, but resolution effects depend on the architecture rather than increasing monotonically.

\begin{table}[!htb]
\centering
\caption{PE-Core-L-14 and PE-Core-G-14 evaluated at the same input resolutions. Task~1 reports MRR with Seed2.0 captions and Task~2 reports nDCG@10. Native resolutions are underlined.}
\label{tab_disc_scale}
\resizebox{\columnwidth}{!}{%
\begin{tabular}{lcccc}
\toprule
& & \multicolumn{2}{c}{\textbf{Task 1 (MRR)}} & \textbf{Task 2} \\
\cmidrule(lr){3-4}\cmidrule(lr){5-5}
\textbf{Backbone} & \textbf{Resolution} & \textbf{T$\to$I} & \textbf{I$\to$T} & \textbf{nDCG@10} \\
\midrule
\multirow{3}{*}{PE-Core-L-14} & 224 & 37.02 & 37.55 & 26.94 \\
 & \underline{336} & 37.50 & 38.76 & 28.94 \\
 & 448 & 37.50 & 38.50 & 28.48 \\
\midrule
\multirow{3}{*}{PE-Core-G-14} & 224 & 20.56 & 19.62 & 7.61 \\
 & 336 & 42.14 & 40.05 & 32.74 \\
 & \underline{448} & 42.08 & 40.61 & 34.48 \\
\bottomrule
\end{tabular}}
\end{table}
\end{revblock}

\makeatletter\setlength{\@fptop}{0pt}\makeatother

\end{document}